\documentclass[12pt]{article}
\usepackage[letterpaper, margin=1in]{geometry}

\usepackage[square,sort,comma,numbers]{natbib}
\usepackage{times}
\usepackage{graphicx}
\usepackage{amsmath}
\usepackage{mathtools}
\usepackage{tabularx}
\usepackage{multirow}
\usepackage{textgreek}
\usepackage{url}

\usepackage{lineno}

\usepackage{color,soul}
\newcommand{\sy}[1]{\textcolor[rgb]{0,0,0}{#1}}

\DeclareMathAlphabet\mathbfcal{OMS}{cmsy}{b}{n}

\newcommand\datum{d}
\newcommand\ts{\tau}

\newcommand\targetlabel{y^*}
\newcommand\contrastivelabel{y}
\newcommand\targetdatum{d^*}

\title{Mitigating belief projection in explainable artificial intelligence via Bayesian Teaching}
\author
{Scott Cheng-Hsin Yang,$^{1\dagger\ast}$, Wai Keen Vong,$^{2\dagger}$ Ravi B. Sojitra,$^{3\dagger}$, \\Tomas Folke$^{1}$, Patrick Shafto$^{1}$ \\
\\
\normalsize{$^{1}$Department of Mathematics and Computer Science, Rutgers University}\\
\normalsize{101 Warren Street, Newark, NJ 07102, USA}\\
\normalsize{$^{2}$Center for Data Science, New York University}\\
\normalsize{60 5th Ave, New York, NY 10011, USA}\\
\normalsize{$^{3}$Department of Management Science and Engineering, Stanford University}\\
\normalsize{$^\dagger$Equal contribution.}
\\
\normalsize{$^\ast$To whom correspondence should be addressed; E-mail: scott.cheng.hsin.yang@gmail.com.}
}

\date{}

\begin{document} 

\maketitle 

\begin{abstract}
State-of-the-art deep-learning systems use decision rules that are challenging for humans to model. Explainable AI (XAI) attempts to improve human understanding but rarely accounts for how people typically reason about unfamiliar agents. We propose explicitly modelling the human explainee via Bayesian Teaching, which evaluates explanations by how much they shift explainees' inferences toward a desired goal. We assess Bayesian Teaching in a binary image classification task across a variety of contexts. Absent intervention, participants predict that the AI's classifications will match their own, but explanations generated by Bayesian Teaching improve their ability to predict the AI's judgements by moving them away from this prior belief. Bayesian Teaching further allows each case to be broken down into sub-examples (here saliency maps). These sub-examples complement whole examples by improving error detection for familiar categories, whereas whole examples help predict correct AI judgements of unfamiliar cases.
\end{abstract}

While Artificial Intelligence (AI) can help address socially-relevant problems \cite{doshi2017accountability,rajpurkar2017chexnet,esteva2017dermatologist}, it is important for humans to be able to scrutinize AI decisions so we may audit, understand, and improve performance; indeed, this is legally mandated in certain contexts \cite{EUdataregulations2018, coyle2020explaining}. \sy{The best performing AI algorithms rely on complex decision rules based on features that feel alien to most humans \cite{lake2017building}. The abstruseness of these AI models impedes their adaptation in high-leverage contexts}, emphasizing the need for successful explanations that facilitate human understanding and prediction of the AI's behavior.

A popular class of methods to explain \sy{AI systems} is \textit{explanation-by-examples}. Explanation-by-examples takes as input an AI model to be explained and the data that it has been trained on and produces as output a small subset of training data that exert high impact on the inference of the explainee. For example, if the aim is to explain whether a deep-learning model would classify a given image as a cat or a dog, explanation-by-examples selects the cat and dog images that are most representative of those categories. The utility of explanation-by-examples is supported by research that confirms humans' ability to induce principles from a few examples \cite{mill1889system,bloom2002children,xu2007word,lake2020people} as well as the extensive use of examples in education \cite{chi1989self,aleven1997teaching,bills2006exemplification}. The explanation-by-examples approach has many desirable properties: It is fully model-agnostic and applicable to all types of machine learning \cite{chen2018learning, eaves2016tractable, ho2016showing}; it is domain- and modality-general \cite{hendricks2018generating, kanehira2019learning}; and it can be used to generate both global explanation \cite{kim2014bayesian, kim2016examples, vong2018, wang2018dataset, koh2017understanding} and local explanation \cite{papernot2018deep, yeh2018representer, goyal2019counterfactual}. Although the technology of explanation-by-examples for XAI has been developed for at least two decades \cite{caruana1999case, keane2019case}, empirical tests and connections to its ecological roots in the social sciences have been limited. 

Explanation-by-examples can be considered a social teaching act, which can be formally captured by \textit{Bayesian Teaching} \cite{yang2017explainable}. In Bayesian Teaching, there are two parties, a teacher (explainer) who selects examples and an \sy{explainee (learner)} who draws inferences. The teacher selects examples intended to maximize the \sy{explainee's} probability of a correct inference based on the teacher's model of the \sy{explainee's} current beliefs and their inductive biases \cite{miller2018explanation, shulman1986those, chick2007teaching}; the \sy{explainee} uses Bayesian updating to make predictions given these examples \cite{Shafto2014,eaves2016infant,eaves2016tractable,yang2018optimal}. Existing work on explanation-by-examples has demonstrated explanation effectiveness relative to several baseline conditions \cite{mac2018teaching, chen2018near, chen2018learning, kim2016examples}; however, there is rarely a principled, \textit{apriori} rationale as to why the proposed improvements should work. By explicating the computations used to model the explainer, the explainee, and the explanation selection process, Bayesian Teaching provides testable predictions on the effectiveness of explanatory examples in different contexts.

\begin{figure}[t]
  \centering
  \includegraphics[width=0.8\columnwidth]{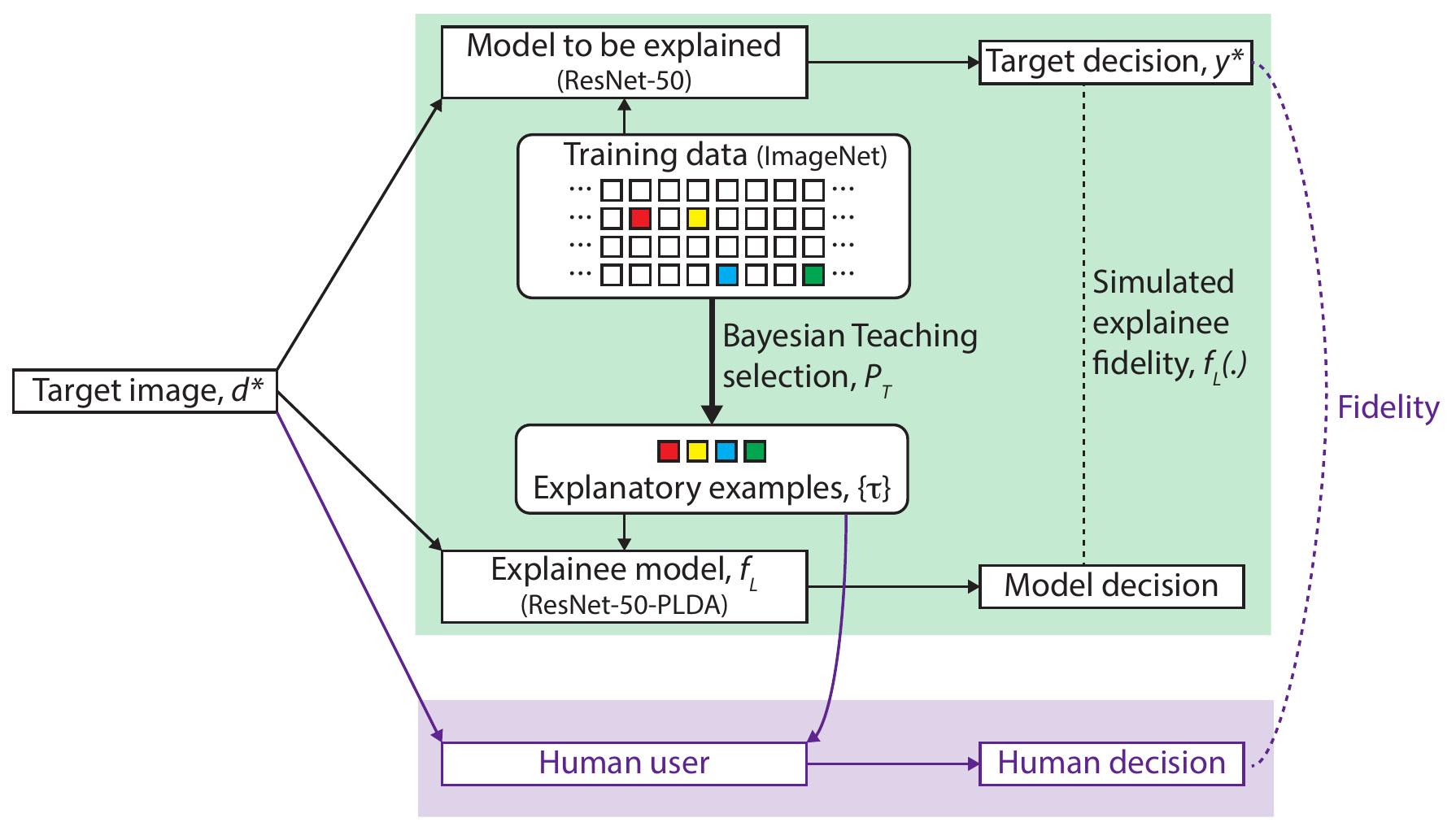}
  \caption{\sy{Overview of Bayesian Teaching. Green section: Bayesian Teaching selects a few examples from the training data as explanatory examples. The explanatory examples are selected such that when the \sy{explainee model} is trained on those examples, the fidelity between the model decision and the target decision (simulated explainee fidelity) matches a desired value. The same logic is applied to generate the saliency maps, where pixels are treated as examples at a finer granularity. Purple section: The explanatory examples and saliency maps are also shown to human participants. Explanation effectiveness is evaluated by how the examples influence the fidelity between participants' decisions and the target decisions. Symbols correspond to those in Equation~\ref{eq:bt_simple}; arrows indicate input-output relationship; and dotted line indicate comparison. The models and data used are indicated in parenthesis. This figure was created using Adobe Illustrator CS6 (v.~16.0.0) \cite{adobeillustrator}.}} 
  \label{fig:overview}
\end{figure}

We use image classification on \textit{the ImageNet 1K dataset} \cite{ILSVRC15} as the testbed. The model to be explained is ResNet-50 \cite{he2016deep}. Following an ideal-observer approach \cite{geisler2003ideal,geisler2011contributions}, we instantiate Bayesian Teaching by selecting examples with differing degrees of helpfulness as judged by the \sy{fidelity} between the \sy{explainee} model and the target model. For the \sy{explainee} model, we used a \sy{ResNet-50-PLDA model, which is a ResNet-50 model} where the last softmax layer is replaced by a probabilistic linear discriminate analysis (PLDA) model. This alteration introduces the probabilistic training required by Bayesian Teaching while keeping the architecture of ResNet-50, which is known to accurately fit human labels \cite{he2016deep}. \sy{In the context of image classification, Bayesian Teaching can be expressed as}
\begin{align}
P_T(\{\tau\}|y^*,d^*) \propto f_L(y^*|d^*,\{\tau\}),
    \label{eq:bt_simple}
\end{align}
\sy{where $d^*$ is a target image; $y^*$ is the label predicted by the model to be explained, hence the target decision; $\{\tau\}$ is a set of explanatory examples; $f_L$ is the \sy{explainee model}; the probability produced by $f_L(\cdot)$ is the simulated \sy{explainee} fidelity; and $P_T(\cdot)$ determines the probability of selecting a set explanatory examples. See Figure~\ref{fig:overview} for an overview and the Methods for further details.}
\sy{Bayesian Teaching also allows for selection of examples at different levels of granularity.} For the current task, we consider the selection of entire images as well as pixels in an image as explanations. The latter pixel-selection process derived from Bayesian Teaching turns out to be mathematically equivalent to a type of feature attribution method called Randomized Input Sampling for Explanations \cite{petsiuk2018rise}. Thus the two levels of example granularity evaluated in this paper coincide with two popular methods of explanation---explanation-by-examples and saliency maps.

\sy{To give a concrete example, consider a trial where a participant tries to predict whether the target AI classifier will classify a certain image as a barn or a flagpole (see Figure 2). Bayesian Teaching operates by selecting the four example images---two of a barn and two of a flagpole---from the training set that are most likely to make the explainee model reach the same judgement as the target model, i.e., high simulated explainee fidelity. The target image and the examples are overlaid with saliency maps where each pixel is weighted by the probability that showing it will guide the explainee model to the same conclusion as the target model.}

\sy{Bayesian Teaching contributes to the literature on XAI by formalizing the role of the explainee. Explicitly considering the explainee highlights how XAI methods can be validated, and how explanations informed by the \sy{explainee model} can mitigate human prior beliefs about the AI system. We showcase three criteria to validate explainable AI from the Bayesian Teaching perspective: (1) Explanations selected by Bayesian Teaching improve the \textit{fidelity} between human prediction of AI classification and actual AI classification; (2) the Bayesian Teacher can correctly infer which explanations humans will prefer; and (3) the Bayesian Teacher can accurately predict both which explanation will improve fidelity and which explanations will decrease it. Additionally, we show how the prior beliefs of human participants can be mitigated by appropriate explanations. Consistent with existing work from psychology \cite{gordon1986folk, koster2013theory}, we find that human participants project their own beliefs onto the AI system. This belief-projection manifests as (4) fidelity being higher when the AI is correct relative to when it is wrong, (5) this impact of AI correctness on fidelity being particularly pronounced for familiar categories, and (6) these effects being mitigated by appropriate explanations. We provide justifications and intuitions for these six points in the following paragraphs. To the best of our knowledge, this is the first paper to empirically explore the implications of human belief-projection for explainable AI.}

\sy{The core prediction of Bayesian Teaching is that explanations which lead the explainee model to correct predictions will help humans to better understand the AI. We test this by evaluating whether participants exposed to helpful examples and saliency maps are better able to predict the AI system's classifications than participants who do not view any explanations. Returning to our example in Figure~\ref{fig:exp}, this means that a participant who is shown the example images and saliency maps are more likely to correctly predict that the AI classified the image as a flagpole rather than a barn, relative to a participant who is only shown the target image without any explanation. This is a generous test of Bayesian Teaching, but a necessary one, because failing this test would make all subsequent results moot. Provided that the explainee model match human users reasonably well, we expect that examples selected to be helpful by the Bayesian Teacher will be preferred over examples that are selected to be unhelpful or at random. A stricter test of the appropriateness of the Bayesian Teaching is whether it can predict both explanations that improve the fidelity of human predictions and those that lead to reduced fidelity. Such calibration implies that it is not every explanation improves fidelity, but that explanations need to be curated to reach a desired result. In our experimental setup this would manifest in examples that are judged to be helpful or detrimental by the Bayesian Teacher increasing or reducing the fidelity of the participants' predictions to the AI judgments, respectively.} 

\sy{If human participants project their beliefs onto the AI system, they will expect the AI classifier to be highly accurate because they themselves perform well at image classification \cite{ILSVRC15}. In our experiment this translates to humans who predict AI classifications achieving higher sensitivity (correctly predicting AI's correct classifications) than specificity (correctly predicting AI’s mistakes), absent explanation. In the context of our example: since the target image is showing a barn, a participant not given any explanation should typically (incorrectly) predict that the AI will classify the image as a barn rather than a flagpole. However, this effect should not be uniform across trials because some categories are easier to distinguish than others. Since more familiar categories should be easier to distinguish, and since participants expect the model to get the right answer for trials they themselves find easy, belief projection implies that \textit{familiarity should increase fidelity for model hits}. Conversely, \textit{familiarity should decrease fidelity for model errors}. Introducing a different example, a participant who is familiar with dogs will find the discrimination between yorkshire terrier and silky terrier easy, whereas someone less familiar with dogs might struggle with the first-order categorization, and consequently be more willing to consider the AI classifier making a mistake.}

\sy{If explanations generated by Bayesian Teaching operates by mitigating belief-projection, we would expect them to reduce the gap between sensitivity and specificity by increasing the latter (improving error detection). Additionally, the belief-projection implies that examples improve fidelity the most for unfamiliar categories, whereas saliency maps improve fidelity most for familiar categories. The reason why examples are most beneficial for unfamiliar categories is that they could strengthen category distinctions for unfamiliar categories with fuzzier mental representations. In the context of the two breeds of terrier: someone who is unfamiliar with dogs can leverage the examples to better understand what features distinguish the two breed, and compare that to the features of the target image. Saliency maps, on the other hand, might be most diagnostic for familiar cases because they highlight features that were consequential to the AI system, and determining the appropriateness of these features requires familiarity with the categories. In the context of the barn versus flagpole example: most people can reliably distinguish between them, so can notice that the saliency map of the target indicates that the AI classifier pays less attention to the house relative to the whethervane, suggesting a potential misclassification.}

\section*{Results}

\subsection*{Methodological overview}
User understanding in the context of classification can be captured by how well the user can predict the model's judgement. Throughout this paper we will refer to this predictive capacity as \sy{fidelity, referring to the agreement between an agent's prediction (either a participant or the \sy{explainee model}) and the judgement of the classifier}. A natural measure of explanation effectiveness is how much the explanations increase such \sy{fidelity}, relative to a control condition. We designed a two-alternative forced choice (2AFC) task in which participants were asked to predict the model's classification of a target image between two given categories. No trial-by-trial feedback was provided to participants. It is important to note that in this task high \sy{fidelity} does not imply that participants' judgements match the ground truth of the image, which we refer to as first-order accuracy or simply accuracy. It is possible for a participant to have high accuracy (in that their judgements often match the ground-truth category of the image) but poor \sy{fidelity} (in that their judgements rarely match the AI's).

\begin{figure}[ht!]
  \centering
  \includegraphics[width=\columnwidth]{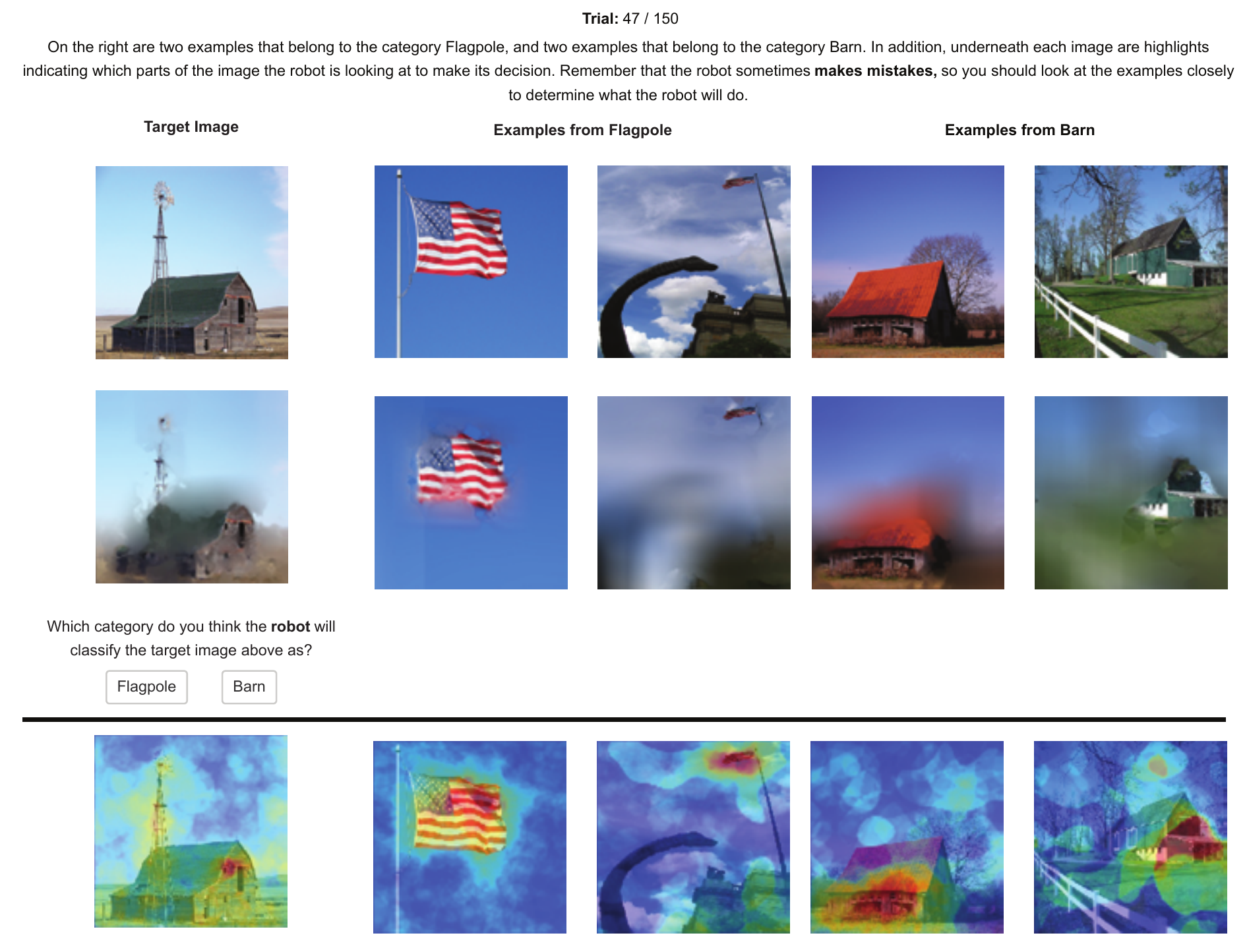}
  \caption{A snapshot of the experiment. See Methods and Table~\ref{tab:participants} for the naming conventions of the conditions used below. The experimental condition above the black line is [\textsc{specific labels}] \& [\textsc{helpful}] \& [\textsc{blur}]. Under the black line is the  [\textsc{jet}] equivalent of the second row, which is obtained by replacing the blurring maps with the jet color maps. Experimental conditions with generic labels are obtained by replacing specific labels---`Flagpole" and ``Barn" in this case---with generic category names---``Category A" and ``Category B." Experimental conditions without the saliency maps, i.e., [\textsc{no map}], show only the first row of images. Conditions without examples, i.e., [\textsc{no examples}], show only the first column of image(s). All images and saliency maps shown were 224-by-224 pixels. The prediction of the model to be explained on the target image is ``Flagpole" in this case. \sy{All photographs are obtained from the open-source \textit{ImageNet 1K} dataset \cite{ILSVRC15}. This figure was created using Adobe Illustrator CS6 (v.~16.0.0) \cite{adobeillustrator}.}} 
  \label{fig:exp}
\end{figure}

We designed a total of 15 conditions that vary along three dimensions: (1) presence of informative labels (two levels: [\textsc{generic labels}] or ([\textsc{specific labels}]), (2) types of examples (three levels: [\textsc{no examples}], [\textsc{helpful}] or [\textsc{random}]), and types of saliency maps (three levels: [\textsc{no map}], [\textsc{jet}] or [\textsc{blur}]). The labels dimension indicate whether the images shown where given informative labels (e.g. "Border terrier" or "Norwich terrier") or generic labels (Category A or Category B). The examples dimension indicate whether examples of the two image categories were shown, and if so, if they were selected to be helpful or were drawn from a uniform distribution of helpfulness as determined by Bayesian Teaching. The saliency map dimension indicates if the images were overlaid with saliency maps that highlighted which pixels the \sy{AI classifier} focused on to make its classification. If saliency maps were included, they were either visualized as a semi-transparent jet color map or as an image filter where unimportant pixels where blurred. We found no significant difference between the [\textsc{blur}] and [\textsc{jet}] conditions; thus, for increased clarity we use the [\textsc{map}] condition, which contains both variants, in the main text. See Supplementary Discussion D2 for the main analyses in the paper repeated with [\textsc{blur}] and [\textsc{jet}] coded separately. Table~\ref{tab:participants} shows the sample size of each condition. Figure~\ref{fig:exp} shows a trial where the categories are represented with informative label, helpful examples, and blur saliency maps.

\begin{table}[ht!]
\centering
\begin{tabularx}{\linewidth}{|c c |c c c c c}
  \cline{3-7}
  \multicolumn{1}{c}{} & &
  \multicolumn{3}{c}{\textsc{specific labels}} &
  \multicolumn{2}{|c|}{\textsc{generic labels}} \\
  \cline{3-7}
  \multicolumn{1}{c}{} & & 
  \multirow{2}{*}{\textsc{no examples}} & 
  \multicolumn{4}{|c|}{\textsc{examples}} \\
  \cline{4-7}
  \multicolumn{1}{c}{} & & & 
  \multicolumn{1}{|c}{\textsc{helpful}} &
  \multicolumn{1}{|c}{\textsc{random}} &
  \multicolumn{1}{|c}{\textsc{helpful}} &
  \multicolumn{1}{|c|}{\textsc{random}} \\
  \cline{1-7}
  \multicolumn{2}{|c|}{\textsc{no map}} &
  \multicolumn{1}{c}{N = 76} &
  \multicolumn{1}{|c}{N = 35} & 
  \multicolumn{1}{|c}{N = 34} &
  \multicolumn{1}{|c}{N = 38} &
  \multicolumn{1}{|c|}{N = 36} \\
  \cline{1-7}
  \multirow{2}{*}{\textsc{map}} &
  \multicolumn{1}{|c|}{\textsc{blur}} &
  \multicolumn{1}{c}{N = 65} &
  \multicolumn{1}{|c}{N = 33} & 
  \multicolumn{1}{|c}{N = 36} &
  \multicolumn{1}{|c}{N = 35} &
  \multicolumn{1}{|c|}{N = 34} \\
  \cline{2-7}
  & \multicolumn{1}{|c|}{\textsc{jet}} &
  \multicolumn{1}{c}{N = 71} &
  \multicolumn{1}{|c}{N = 33} & 
  \multicolumn{1}{|c}{N = 35} &
  \multicolumn{1}{|c}{N = 35} &
  \multicolumn{1}{|c|}{N = 35} \\
  \cline{1-7}
\end{tabularx}
\caption{Naming convention of conditions and the number of participants in each condition. In the main text, conditions are referred to with brackets and the ``\&" logical operation. For example, [\textsc{no examples}] \& [\textsc{no map}] refers to the condition with 76 participants (See Experimental conditions in Methods for more detail).}
\label{tab:participants}
\end{table}

Each trial has three more distinct features beyond the condition it belongs to: the \sy{category accuracy}, the \sy{simulated explainee fidelity}, and a familiarity score. \sy{Category accuracy} refers to the classification accuracy on the category which the target ResNet-50 model predicts that the target image belongs to (see Supplementary Table T1). Note that in contrast to the \textit{\sy{category accuracy}} which is an accuracy on the category-level, we use the term \textit{model correctness} to refer to whether the target model made a correct judgement on a specific trial. The \sy{simulated explainee fidelity} of a trial (only available in the [\textsc{examples}] conditions) is an estimate of the probability that the \sy{explainee model}'s classification would match the target ResNet-50 model's classification, given the categories and examples presented. Finally, in a separate study seven raters indicated their familiarity with each category pairing by stating whether they thought they could correctly match images of the two categories presented to their respective labels. The familiarity score is the mean value across all seven raters. See the Methods for a more technical explanation of these features.

\subsection*{Bayesian Teaching improves \sy{fidelity}}
\begin{figure}[ht!]
  \centering
  \includegraphics[width=\columnwidth]{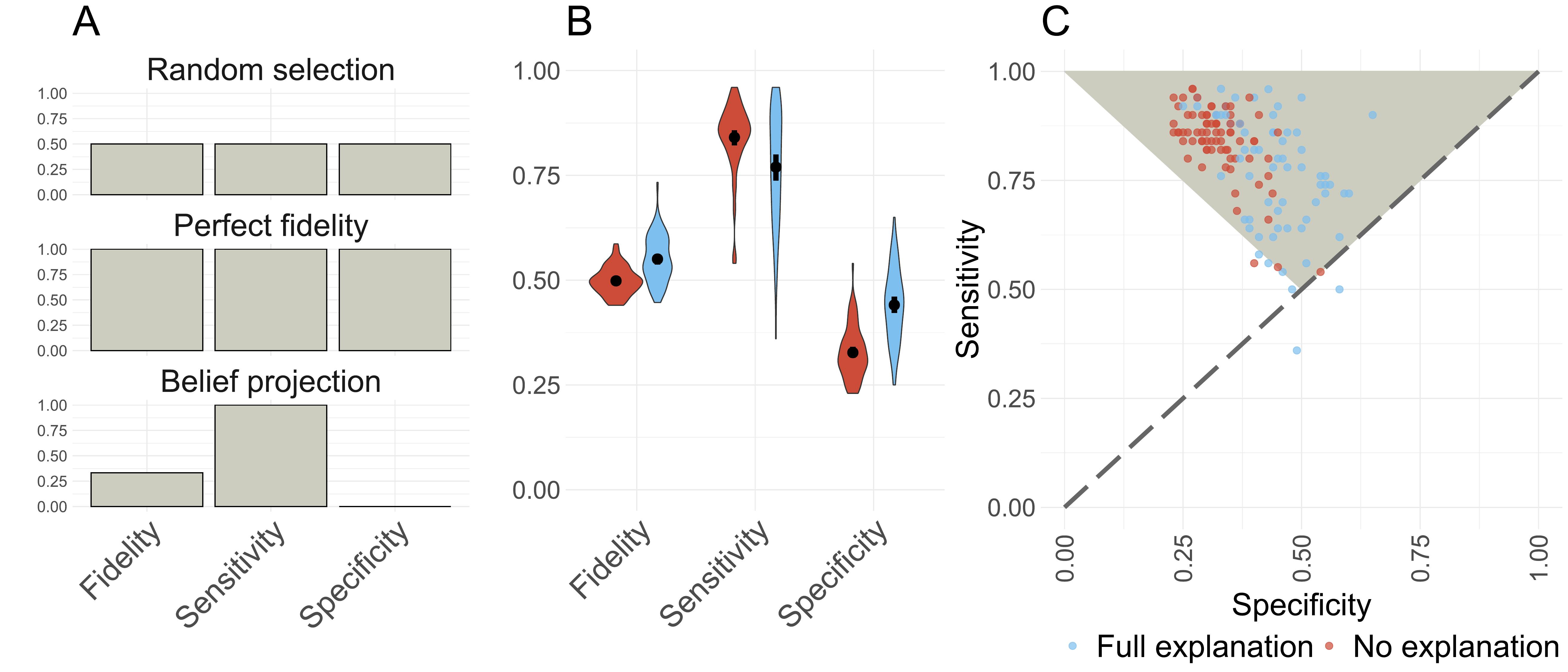}
  \caption{\sy{Bayesian Teaching improves fidelity by mitigating belief projection.} The effectiveness of examples generated by Bayesian Teaching, evaluated by comparing the \sy{fidelity} of the participants who obtained a full explanation ([\textsc{specific labels}] \& [\textsc{helpful}] \& [\textsc{map}]; 66 participants; 9,899 observations) with a control group ([\textsc{specific labels}] \& [\textsc{no examples}] \& [\textsc{no map}]; 76 participants; 11,394 observations). \textbf{(A)}. Three idealised \sy{fidelity} profiles, showing the \sy{fidelity} of: a random agent, a perfect agent, and an agent with perfect access to the ground truth who assumes that the \sy{AI classifier} always mirror their own predictions (belief projection). \textbf{(B)}. Human \sy{fidelity} most closely match the belief projection profile, but the interventions increase specificity (and slightly reduce sensitivity) by making participants better at spotting the AI's errors. The violinplots show the distribution of \sy{fidelity} within conditions. Black dots show the group mean with error bars signifying 95\% bootstrapped confidence intervals. \textbf{(C)}. Individual participants' sensitivity and specificity. The vertices of the triangle show the \sy{fidelity} of a belief-projecting agent with perfect access to the ground truth (upper left), an agent with a perfect model of the \sy{AI classifier} (upper right), and an agent choosing at random (lower middle).  The control group is clustered at high sensitivity and low specificity towards the upper left, whereas the experimental group is shifted to the right. However, the experimental group also shows greater variance, signifying inter-individual differences in the intervention effectiveness. \sy{This figure was created using the ggplot2 package (v. 3.3.2) \cite{ggplot} in R (v. 4.0.3) \cite{r_core}.}}
  \label{fig:exp_eff}
\end{figure}

To evaluate whether the XAI interventions improved \sy{fidelity} we compared participants who obtained a full explanation ([\textsc{specific labels}] \& [\textsc{helpful}] \& [\textsc{map}]) with a control group that received no explanations ([\textsc{specific labels}] \& [\textsc{no examples}] \& [\textsc{no map}]). When interpreting these results in relation to belief projection it is instructive to consider three idealized scenarios. An agent who picked categories at random would have 50\% \sy{fidelity}, sensitivity (correctly predicting AI classifications when the \sy{AI classifier} is correct), and specificity (correctly predicting the AI's mistakes). An agent who modelled the \sy{AI classifier} perfectly would have 100\% \sy{fidelity}, sensitivity, and specificity. Finally, an agent with perfect first-order accuracy who projected their own beliefs onto the \sy{AI classifier} would have 100\% sensitivity, 0\% specificity, and 33\% overall \sy{fidelity} because the experiment contains twice as many AI errors as AI correct classifications (see Methods). Absent intervention, participants behave most like the third, belief-projecting, agent (Figure~\ref{fig:exp_eff}). 

The explanation interventions increase overall \sy{fidelity} by increasing specificity (participants are better able to spot the AI's mistakes), at the cost of some sensitivity. Participants in the control condition have a mean \sy{fidelity} of 49.83\% [95\% CI = 48.83\% - 50.84\%], significantly lower than the 55.04\% [95\% CI = 52.58\% - 57.48\%] \sy{fidelity} of the experimental group (\textbeta = 0.21(0.03), z = 6.99, p $<$ .0001). This is primarily driven by higher specificity in the experimental group (43.98\% [95\% CI = 39.68\% - 48.37\%] relative to the control group's 32.54\% [95\% CI = 30.96\% - 34.13\%]; \textbeta = 0.49(0.05), z = 9.20, p $<$ .0001). The greater vigilance of the experimental group came with a minor cost to sensitivity for the experimental group (78.90\% [95\% CI = 71.59\% - 84.80\%]) and  for the control group (85.26\% [95\% CI = 83.12\% - 87.22\%]); \textbeta = -0.43(0.12), z = -3.68, p = .0002), but not enough to offset the specificity gains. Collectively, these results imply that participants attempt to predict the AI by projecting their own beliefs, and that the explanations improve \sy{fidelity} by mitigating this belief projection. 

\subsection*{Participants prefer examples that are helpful according to Bayesian Teaching}
Having established that examples generated by Bayesian Teaching improved participants’ ability to predict AI judgements, we want to evaluate whether participants preferred helpful to random and misleading examples. To test this, we ran a second study where participants chose between helpful examples versus random examples or versus misleading examples, where helpfulness was determined by Bayesian Teaching. Participants showed a small but reliable preference for helpful relative to random examples and a substantial preference for helpful versus misleading examples. \sy{Consistent with our hypothesis that helpful examples are most beneficial for unfamiliar categories}, our results show that the preference for helpful examples was particularly pronounced when the image categories were unfamiliar (see Supplementary Discussion D1 for all the details).

\subsection*{Bayesian Teaching can \sy{predict which explanations improve and reduce fidelity}}

Bayesian Teaching makes explicit the existence of an explainee and suggests that a sound \sy{explainee model} should have the capacity to track the inference of actual explainees. In our experiment the calibration between the \sy{explainee model} and the participants is captured by the relationship between \sy{category accuracy} and participant accuracy. We estimate participant accuracy (their first-order belief about the ground truth) by using their \sy{fidelity} in the control trials (their second-order belief about the \sy{AI classifier} with no exposure to explanation). The assumption that their attempt to predict the \sy{AI classifier} may serve as a proxy of their first-order accuracy is justified given the tendency to belief-project observed in previous sections. We found that participant \sy{fidelity} (interpreted as accuracy for the control trials) was positively correlated with \sy{category accuracy} for trials where the model was correct (\textbeta = 1.74(0.20), z = 8.67, p $<$ .0001), indicating good calibration between the model and participants in this situation (see Supplementary Figure F1). We also found a negative interaction between \sy{category accuracy} and model correctness (\textbeta = -2.57(0.23), z = -11.03, p $<$ .0001). This suggests the poor calibration in the special case in which the model's overall accuracy on the predicted category is high but it misclassifies the particular trial. \sy{In sum, these results imply that \sy{category accuracy} is a good proxy of human ground truth judgements at the aggregate level, which in turn suggests that our explainee model is appropriate for our participants.}

Bayesian Teaching should be able to modify participant \sy{fidelity} by selecting explanations of varying helpfulness. To test this in practice, we ran three nested hierarchical logistic regression models of increasing complexity. Each regression model predicted participant \sy{fidelity} (whether the participant correctly predicted the \sy{AI classifier} on a given trial) from the [\textsc{examples}] trials only, as these are the only trials impacted by the \sy{simulated explainee fidelity}, which measures the degree to which the examples would lead the \sy{explainee model} to the targeted inference. The first regression model \sy{served as a null-model, not using simulated explainee fidelity as a predictor, only including \sy{category accuracy} and a dummy variable encoding AI correctness (whether the AI prediction for that trial matched the ground truth or not)}. The second regression model \sy{added simulated explainee fidelity as a predictor, capturing the hypothesis that the helpfulness of the examples as determined by Bayesian Teaching covaries with participant fidelity.} The third regression model \sy{added two two-way interactions between model correctness (model hit and error) and \sy{category accuracy}, and model correctness and simulated explainee fidelity, capturing the hypothesis that helpful examples had differential impact on error detection relative to hit confirmation}. We found that the second regression model fitted the \sy{fidelity} data better than the first regression model (\textchi2(1, 4) = 71.68, p $<$ .0001). This means that \sy{the Bayesian Teacher's perception of the helpfulness of the presented examples} predict participant \sy{fidelity} above and beyond \sy{category accuracy}. The third regression model outperformed the second regression model (\textchi2(3, 7) = 7371.28, p $<$ .0001). This indicates that how well \sy{the category accuracy and/or the modelled helpfulness of the examples shown} predicted fidelity differed for trials with correct or incorrect AI judgements.

\begin{figure}[ht!]
  \centering
  \includegraphics[width=\columnwidth]{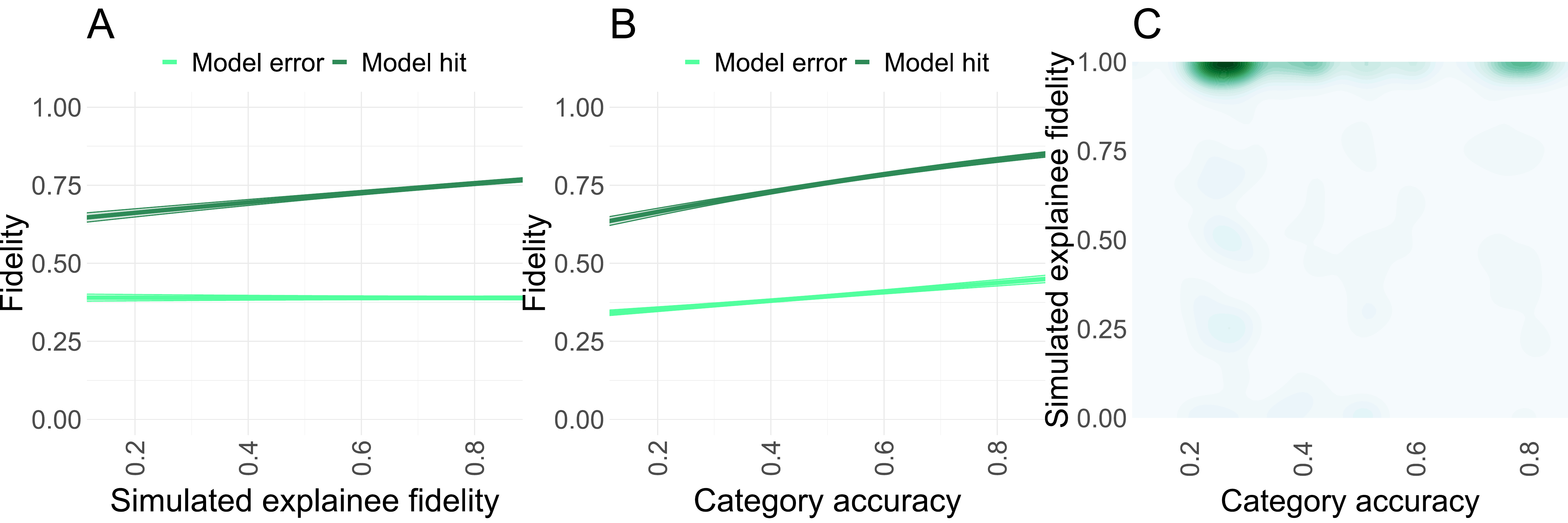}
  \caption{\sy{The helpfulness of the presented examples as determined by the Bayesian Teacher} predicts human \sy{fidelity} across trials with examples (419 participants; 62,820 observations). \textbf{(A)}. The \sy{simulated explainee fidelity}---the helpfulness of the explanatory examples expected by the Bayesian Teacher---correlates significantly with participant \sy{fidelity} for correct trials but not for incorrect trials. This suggests that the Bayesian Teaching framework can \sy{predict} explanations that are informative or misleading for trials that are correctly classified by the model, but not for trials that are incorrectly classified. \textbf{(B)} \sy{Category accuracy} is positively associated with participant \sy{fidelity}, both for trials when the \sy{AI classifier} is correct and when it is wrong. A similar trend is observed in the control condition (see Supplementary Figure F1). This suggests that humans and ResNet-50 find the same categories difficult to discriminate, \sy{implying that the ResNet architecture can serve as an appropriate model of human participants in this task}. The difference in \sy{fidelity} between when the \sy{AI classifier} is correct and when the \sy{AI classifier} is wrong suggests that it is harder to teach incorrect judgements, at least in this context. \textbf{(C)}. Two-dimensional kernel density with 25 density bins showing the distribution of trials in terms of \sy{category accuracy} and \sy{simulated explainee fidelity}. In this study the two are independent. Note that the higher density near perfect \sy{simulated explainee fidelity} was due to all the helpful examples being selected based on this variable, so they constitute a majority of our example trials. \sy{This figure was created using the ggplot2 package (v. 3.3.2) \cite{ggplot} in R (v. 4.0.3) \cite{r_core}.}} 
  \label{fig:teacher_model}
\end{figure}

To explore how model correctness interacted with \sy{category accuracy} and \sy{simulated explainee fidelity}, we explored the parameters of the third regression model. Participants are typically better at predicting the \sy{AI classifier} when it is correct relative to when it is wrong (\textbeta = 0.53(0.06), z = 9.15, p $<$ .0001). This aligns with our previous results, which suggest that participants have a sense of the ground truth for most trials, and assume that the \sy{AI classifier} would make the same judgement that they would make. \sy{Category accuracy} is positively associated with participant \sy{fidelity} when the AI is wrong (\textbeta = 0.59(0.05), z = 12.30, p $<$ .0001), and even more strongly associated with fidelity when the \sy{AI classifier} is correct (\textbeta = 0.93(0.09), z = 10.68, p $<$ .0001; see Figure~\ref{fig:teacher_model}). Because there was a significant positive relationship between ResNet \sy{accuracy and participant fidelity} for both the control trials and the example trials, it seems plausible that the calibration between model and participant observed in the control condition \sy{survives the introduction of explanatory examples}, at least partially. Finally, while statistically controlling for \sy{category accuracy}, \sy{simulated explainee fidelity} did not predict \sy{fidelity} on trials when the \sy{AI classifier} was wrong (\textbeta = -0.01(0.03), z = -0.16, p = .89) but did so for trials when the \sy{AI classifier} was correct (\textbeta = 0.77(0.05), z = 14.19, p $<$ .0001). Because the \sy{simulated explainee fidelity} determined which examples were shown, the fact that this variable could accurately predict human \sy{fidelity} above and beyond ResNet \sy{accuracy} implies that the Bayesian Teacher can \sy{successfully predict which explanations improve \textit{or impair} the fidelity of} participant judgements.

\subsection*{Bayesian Teaching improves \sy{fidelity} through belief-mitigation}
The previous results indicate that examples \sy{deemed helpful by the Bayesian Teacher} improve participant predictions of the \sy{AI classifier's judgements}. Additionally, participants prefer examples that are helpful according to the Bayesian Teacher, and this preference is particularly pronounced for unfamiliar categories. Next, we will explore \textit{how} explanatory examples improve \sy{fidelity}, and evaluate the relative importance of the different explanation features employed. The preceding results imply that people belief-project by default: that is, they use their own beliefs as priors for the \sy{AI classifier's} beliefs. The interventions shift these priors, allowing the participants to distinguish their first-order beliefs about the correct classification from their second-order beliefs about the \sy{decisions} of the \sy{AI classifier}.

To further evaluate whether explanations improve \sy{fidelity} by mitigating belief-projection, we compared how the interventions impacted \sy{fidelity} and first-order accuracy in the complete data set. Specifically we contrasted [\textsc{specific labels}] vs [\textsc{generic labels}], [\textsc{map}] vs [\textsc{no map}], and [\textsc{examples}] vs [\textsc{no examples}], while controlling for \sy{category accuracy} and familiarity score. We ran separate analyses for when the \sy{AI classifier} was correct and when the \sy{AI classifier} was wrong, corresponding to the distinction between sensitivity and specificity in previous sections. We will treat the ground truth as a proxy of participant first-order beliefs, \sy{a} defensible assumption given the reported human accuracy on ImageNet in previous works \cite{ILSVRC15}. Based on this assumption, interventions increasing \sy{fidelity while also increasing mismatches to the ground-truth}, would \sy{shift participant predictions of the AI classifier} away from their first-order judgements. The [\textsc{specific labels}] are associated with higher \sy{fidelity} than the [\textsc{generic labels}] regardless of whether the \sy{AI classifier} is correct (\textbeta = 0.24(0.08), z = 3.06, p = .002) or not (\textbeta = 0.07(0.03), z = 2.13, p = .03). Because these effects are small and orthogonal to belief projection, they will not be discussed further.

\begin{figure}[h!]
  \centering
  \includegraphics[width=0.9\columnwidth]{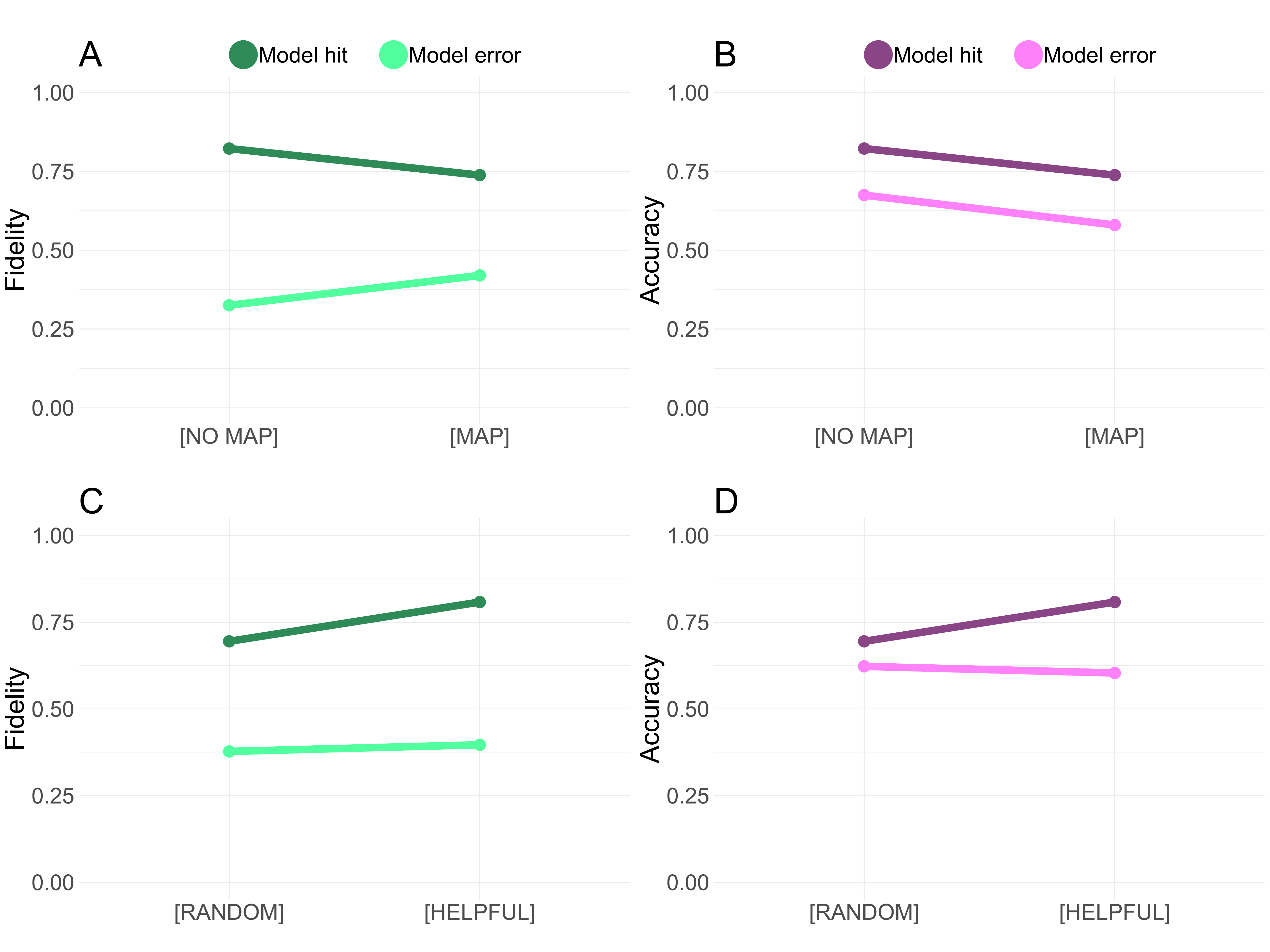}
  \caption{\sy{The fidelity between the participant predictions and the AI classifications is higher when the AI is correct than when the AI is wrong.} \textbf{(A)} \& \textbf{(B)} are based on the entire data set, comparing all [\textsc{map}] conditions  to all [\textsc{no map}] conditions (631 participants; 94,582 observations). \textbf{(C)} \& \textbf{(D)} exclude the [\textsc{no examples}] trials and contrast all [\textsc{helpful}] trials with all [\textsc{random}] trials (419 participants; 62,820 observations). \textbf{(A)}. The saliency maps improve \sy{fidelity} for trials when the \sy{AI classifier} is wrong but reduce \sy{fidelity} when the \sy{AI classifier} is correct. \textbf{(B)}. The saliency maps make people less likely to \sy{predict that the AI classification of the target image matches the ground truth}. Together, (A) \& (B) imply that the saliency maps help people to consider that the \sy{AI classifier} might make mistakes. \textbf{(C)} In trials with examples, helpful examples tend to help people accurately model the \sy{AI classifier} in cases when the \sy{AI classifier} is correct, but have a limited impact when the \sy{AI classifier} is wrong. \textbf{(D)} Consequently, helpful examples make participants more likely to pick the ground truth option when the \sy{AI classifier} is correct, but do not really impact the probability of selecting the ground truth option when the \sy{AI classifier} is wrong. Collectively, these results suggest that helpful examples and saliency maps improve human understanding of the \sy{AI classifier} in distinct and complementary ways\sy{: saliency maps improve error detection, whereas helpful examples enable participants to accurately determine when the AI classifier is correct}. Errorbars represent 95\% bootstrapped confidence intervals. All point estimates have confidence intervals, though some are too narrow to see clearly. \sy{This figure was created using the ggplot2 package (v. 3.3.2) \cite{ggplot} in R (v. 4.0.3) \cite{r_core}.}} 
  \label{fig:intervention_effectiveness}
\end{figure}

The presence of the saliency maps in the [\textsc{map}] condition improves \sy{fidelity} when the \sy{AI classifier} is wrong (\textbeta = 0.43(0.03), z = 14.24, p $<$ .0001), but reduces \sy{fidelity} (to a lesser extent) when the \sy{AI classifier} is correct (\textbeta = -0.56(0.07), z = -7.98, p $<$ .0001; see Figure~\ref{fig:intervention_effectiveness}). In both cases, saliency maps reduced the first order-accuracy of the participants (model hit: \textbeta = -0.56(0.07), z = -7.98, p $<$ .0001; model error: \textbeta = -0.43(0.03), z = -14.24, p $<$ .0001), meaning that they were less likely to report that the \sy{AI classifier}'s judgements matches the ground truth of the image. This implies that the saliency maps encourage participants to consider that the \sy{AI classifier} might be mistaken. One potential explanation for this observation is that the saliency maps show when the \sy{AI classifier} attends to non-sensible features (i.e. parts that are not representative of either of the categories) as well as ambiguous features (e.g. thin metal strips that are present in both the “Electric Fan” and “Buckle” category). 

Comparing all [\textsc{examples}] trials to all [\textsc{no examples}] trials, the presence of examples do not significantly improve \sy{fidelity} when the \sy{AI classifier} is correct (\textbeta = -0.13(0.08), z = -1.61, p = .11) or when the \sy{AI classifier} is wrong (\textbeta = 0.02(0.03), z = 0.69, p = .49). However, in the conditions where examples were present, helpful examples improve \sy{fidelity} for trials when the \sy{AI classifier} was correct (\textbeta = 0.77(0.08), z = 10.11, p $<$ .0001), but not for trials when the \sy{AI classifier} was wrong (\textbeta = 0.06(0.04), z = 1.77, p = .08). The positive influence of the helpful but not the random examples \sy{illustrates that it is not the mere presence of examples that improves fidelity, but that examples have to be carefully selected to be beneficial. Note also that the effect of helpful examples is the opposite} to what we found for the saliency maps: Whereas saliency maps help participants to identify trials when the \sy{AI classifier} has made a mistake by exposing \sy{inappropriate} sub-image-level features, the examples help reinforce participant's prior beliefs for trials in which the \sy{AI classifier} is correct (see Figure~\ref{fig:intervention_effectiveness}). In other words, the saliency maps and the examples serve separate and complementary functions in explaining AI judgements to the participants.

\begin{figure}[ht!]
  \centering
  \includegraphics[width=0.9\columnwidth]{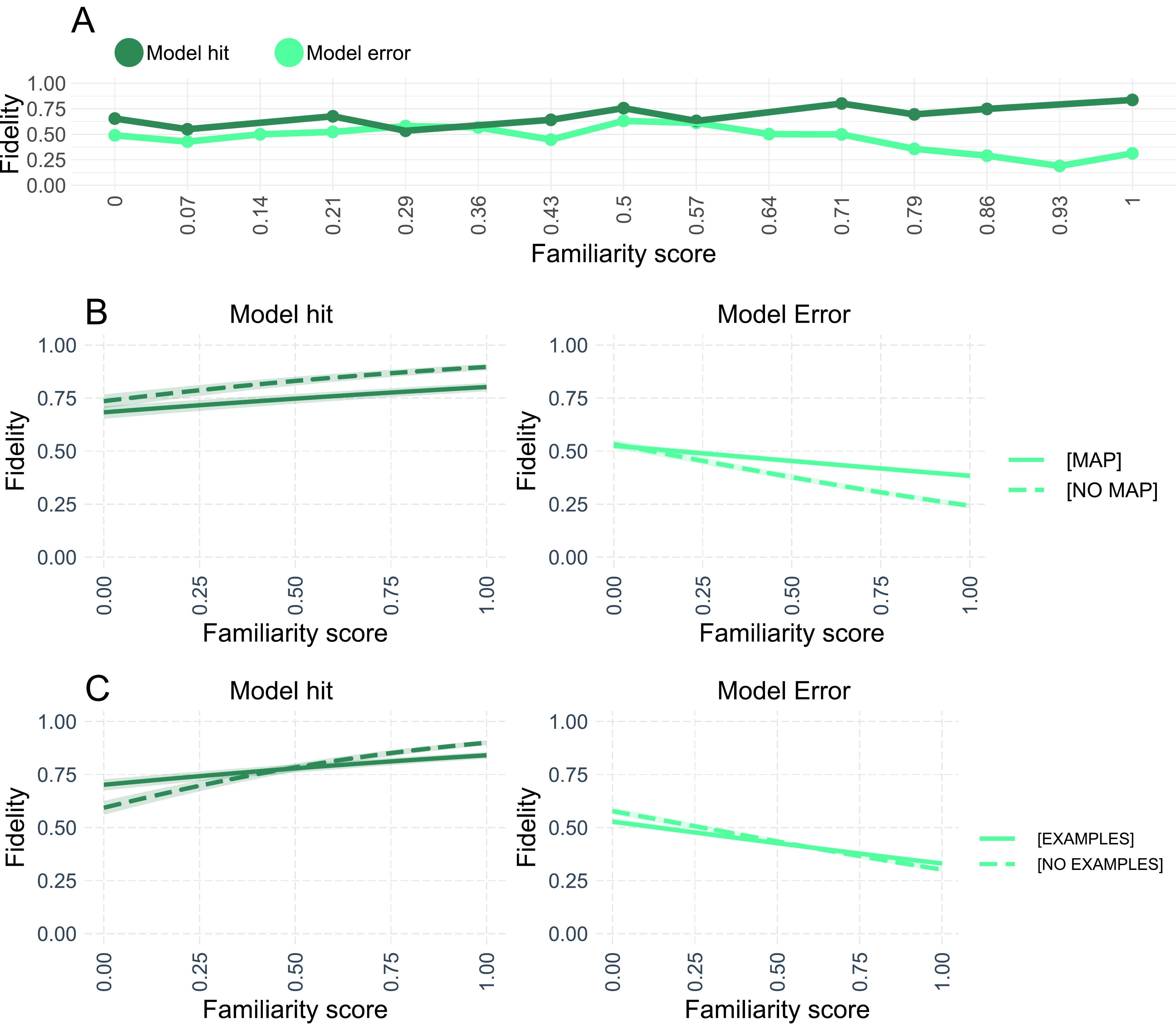}
  \caption{Familiarity score predicts \sy{fidelity} based on the full data set (631 participants; 94,582 observations). \textbf{(A)}. \sy{The fidelity between the participant predictions and the AI classifier's judgements increase with category familiarity when the AI is correct, but decrease with familiarity when it is wrong.} This provides further evidence that participants project their own beliefs \sy{unto the AI classifier, as they are more likely to predict that the AI makes the correct choice on trials they themselves find easy}. \textbf{(B)} Saliency maps decrease the impact of familiarity on participant judgements. For model hits this leads to decreased \sy{fidelity}, whereas for model errors it leads to improved \sy{fidelity}.  This pattern provides further evidence that the saliency maps work by shifting participants away from using their first-order judgments to model the AI's classifications. \textbf{(C)} Examples also decrease the impact of familiarity on participant judgements. For model hits this improves \sy{fidelity} for unfamiliar items but decreases \sy{fidelity} for familiar items, with the opposite pattern for model errors. These results suggest that examples are most beneficial for unfamiliar items when the \sy{AI classifier} is correct. Errorbars represent 95\% bootstrapped confidence intervals. All point estimates have confidence intervals, though some are too narrow to see clearly. Shaded areas represent analytic 95\% confidence intervals. \sy{This figure was created using the ggplot2 package (v. 3.3.2) \cite{ggplot} in R (v. 4.0.3) \cite{r_core}.}} 
  \label{fig:familiarity}
\end{figure}

The familiarity scores capture the ease of the discrimination task in that they are higher for trials involving categories that humans are familiar with. These scores provide clues as to whether participants project their own beliefs onto the AI: If humans use their first-order classifications to model the AI, participants should assume that the \sy{AI classifier} gets the correct answer for trials that they themselves find easy. This is indeed what we find: familiarity is positively associated  with \sy{fidelity} when the and the \sy{AI classifier} is correct (\textbeta = 1.10(0.04), z = 29.28, p $<$ .0001), but negatively associated with \sy{fidelity} for AI errors (\textbeta = -0.92(0.02), z = -42.82, p $<$ .0001; Figure~\ref{fig:familiarity}).

Previously, we showed that saliency maps improved \sy{fidelity} on trials when the \sy{AI classifier} was wrong. This could be explained by saliency maps helping participants distinguish between their first-order judgements of the ground truth and their second-order beliefs about the model classification. This explanation can be evaluated by testing whether the impact of the familiarity scores on \sy{fidelity} are attenuated by the saliency maps. In other words, if participants are more likely to predict that the \sy{AI classifier} is correct on trials that they themselves find easy, and the saliency maps work by helping people realize that the \sy{AI classifier} use decision-processes that differ from their own, the saliency maps should make participants more willing to consider that the \sy{AI classifier} might be wrong for trials they themselves find easy. This is what we find (see Figure~\ref{fig:familiarity}): the presence of saliency maps reduces the positive impact of familiarity on \sy{fidelity} when the \sy{AI classifier} is correct (\textbeta = -0.51(0.08), z = -6.31, p $<$ .0001). Conversely, saliency maps reduce the negative impact of familiarity on \sy{fidelity} when the AI is wrong (\textbeta = 0.70(0.05), z = 15.22, p $<$ .0001; Figure~\ref{fig:familiarity}). Collectively these results suggest that the presence of saliency maps helps participants model the AI as an agent with distinct beliefs that may conflict with their own.

Though the presence of examples did not generally impact \sy{fidelity}, it is possible that they impacted judgements specifically for unfamiliar categories. Like the saliency maps, examples typically reduced the impact of familiarity on \sy{fidelity}, both when the \sy{AI classifier} is correct (\textbeta = -1.01(0.08), z = -12.71, p $<$ .0001) and when the \sy{AI classifier} is wrong (\textbeta = 0.33(0.05), z = 7.35, p $<$ .0001). However, in contrast to the saliency maps, examples seem to be most helpful for unfamiliar trials when the \sy{AI classifier} is correct, see Figure~\ref{fig:familiarity} C. This effect may imply that the examples help participants develop a working representation of the unfamiliar categories, which they are otherwise lacking. 

\section*{Discussion}
\sy{Bayesian Teaching provides a novel way to think about XAI by explicitly modeling the explainee and their prior beliefs. It suggests that explanations can be evaluated in terms of how well they shift explainees' beliefs away from their prior towards a target. We have presented evidence that a Bayesian Teacher can successfully predict which explanations will improve the fidelity between human predictions and target classifications as well as be preferred by human users. Crucially, our results show that the Bayesian Teacher is well-calibrated to human users: it both knows which explanations will improve predictions about the AI and which explanations are problematic. This calibration provides strong evidence that the selection process of the Bayesian Teacher has a causal effect on explainee understanding. Not all examples are created equal, so they need to be appropriately curated.}

\sy{Multiple strands of evidence from our results suggest that in the absence of explanations people project their own first-order beliefs onto the \sy{AI classifier}}. Specifically, we find that participants in the control condition show higher sensitivity than specificity, and that this discrepancy becomes more extreme the more familiar participants are with the trial categories. \sy{The finding that participants predict the AI system by projecting their own beliefs onto the AI links research on explainable AI to the rich psychological literature on social prediction. In many social prediction tasks (in contrast to mechanistic prediction tasks) people use their own preferences, judgements, and beliefs as priors for other agents \cite{tarantola2017prior, suzuki2016behavioral, koster2013theory, bio2018projecting, gordon1986folk}.} Our results imply that such belief-projection can be mitigated by Bayesian Teaching. The most compelling evidence that explanations mitigate belief projection is that the impact of familiarity on \sy{fidelity} is reduced by explanations: explanations make participants more likely to catch AI mistakes on trials they themselves found easy.  

Bayesian Teaching also gives a coherent framework for comparing and contrasting explanatory methods that hereto have been considered independent: explanation-by-examples and feature attribution. We apply Bayesian Teaching to study explanation-by-examples, a popular method for XAI that previously has lacked a sound theoretical footing. Explanation-by-examples has many strengths: it is model-agnostic, domain-general, and easy to use with other XAI methods. Viewed through a Bayesian Teaching lens, this method can be generalized to include feature attribution, another popular post-hoc method, by splitting each example into its component features (i.e. pixels in this study) and considering each pixel individually. When applied to images, such feature attribution on the pixel level generates saliency maps, which is arguably the most popular method for XAI in the image domain. The connection between feature attribution and pixel selection by Bayesian Teaching opens up the possibility to reinterpret all feature attribution methods (e.g., \cite{ribeiro2016should,lundberg2017unified}) as a form of teaching. By treating images and saliency maps as explanatory examples at different levels of granularity, we discover that the two explanations show complementary effects. Namely, example images are effective explanations for confirming the model’s correct classification of unfamiliar categories, and saliency maps are effective explanations for exposing the model’s incorrect classification of familiar categories. 

The lack of a coherent theory is currently stifling XAI as methods are developed around technical innovations without any \textit{apriori} hypothesis as to whether they are appropriate for the specific use case \cite{doshi2017towards}. Bayesian Teaching both exposes this blind spot and offers a solution: effective explanation is a communication act which depends on a knowledgeable teacher, a good model of the explainee, and an awareness of the context in which inference takes place. Consequently, the framework encourages systematic evaluation of XAI interventions on these dimensions, and provides a way to systematically diagnose how interventions could be improved. In our study we show how such an evaluation applies to explanation-by-examples. We modeled the explainee by a ResNet-50 architecture, focused on two contextual variables (familiarity and model correctness), and surfaced \sy{how explanations generated by Bayesian Teaching can mitigate mistaken prior beliefs}. These results highlight the promise of the Bayesian Teaching approach, since the function of explanation is to shape the explainee's inductive reasoning \cite{lombrozo2006structure}. Furthermore, Bayesian Teaching exemplifies how XAI can be improved by considering links to other fields such as education and cognitive science. A balanced synergy between the social sciences and the more technical literature of AI is much needed, as XAI is simultaneously a machine-learning problem and a human-centered endeavor. 

\section*{Methods}
The objective of this study was to explore the effects of explanations, in the form of examples and saliency maps, on users' understanding of  high-performing machine learning models (referred to as AI throughout the paper) in the domain of image classification. We probe users' understanding by a two-alternative-forced-choice (2AFC) task in which users are asked to predict the model's classification of a target image into one of two categories. Experimental conditions vary in terms of the information presented on the screen during each classification. The information presented differs along three dimensions: types of labels, types of examples, and types of saliency maps. All the examples and saliency maps are generated by the Bayesian Teaching framework. The \sy{fidelity} of the participant is captured by sensitivity, specificity, and accuracy.

\subsection*{The model to be explained} 
The machine learning model to be explained is a ResNet-50 model \cite{he2016deep}. For this study, we used the pre-trained version of ResNet-50 in Keras with ImageNet weights. For the selection of saliency maps, the Bayesian Teaching framework expects the model to be able to make probabilistic inference on the image classification task presented in the ImageNet Large Scale Visual Recognition Challenge 2012 (ILSVRC2012). The ResNet-50 model has this capability, and we can use the ResNet-50 model without any modification. However, for the selection of examples, the Bayesian Teaching framework expects the model to be able to make probabilistic inference on the 2AFC task, and the ResNet-50 model is deterministic. We replace the fully-connected classification layer of the ResNet-50 model with a probabilistic linear discriminant analysis (PLDA) model \cite{ioffe2006probabilistic}. This new PLDA layer is trained using a transfer-learning-like procedure. Training images were first passed through the ResNet-50 model and transformed into feature vectors. Then, the PLDA layer was fit to these feature vectors and the corresponding class labels following the algorithm presented in \cite{ioffe2006probabilistic}. Using the training dataset \textit{ImageNet 1K} from the ILSVRC2012 \cite{ILSVRC15}, this ResNet-50-PLDA model has a top-1 accuracy of 52.86\% and a top-5 accuracy of 76.29\%. For the actual experiment, we focused on a subset of 100 categories that include the most difficult, easiest, and most confusable categories (see the next subsection for details). Unless otherwise stated, all the model predictions used to design the experiment is based on the ResNet-50-PLDA model trained on the training data in only these 100 categories.

\subsection*{Stimuli selection}
Each experiment consisted of 150 trials. For 50 of the trials, the predictions of the model (or the robot) matched the ground-truth labels of the target images. For the remaining 100, the model predictions did not match the ground-truth labels. We selected the target images and the classification categories based on the model's confusion matrix, with the aim to cover a wide range of model behavior. First we calculated the ResNet-50-PLDA model's confusion matrix on \textit{ImageNet 1K}, which contains 1000 categories. Then, we randomly selected 25 categories from each of the following four subsets: the 100 categories on which the model was most accurate, the 100 categories that were most confusable with these most accurate categories, the 100 categories on which the model was least accurate, and the 100 categories that were most confusable with these least accurate categories. This resulted in 100 categories. We recorded the model's predicted labels of all the training images in these 100 categories and marked all images for which the model predictions were also among these 100 categories. 

\begin{figure}[ht!]
  \centering
  \includegraphics[width=\columnwidth]{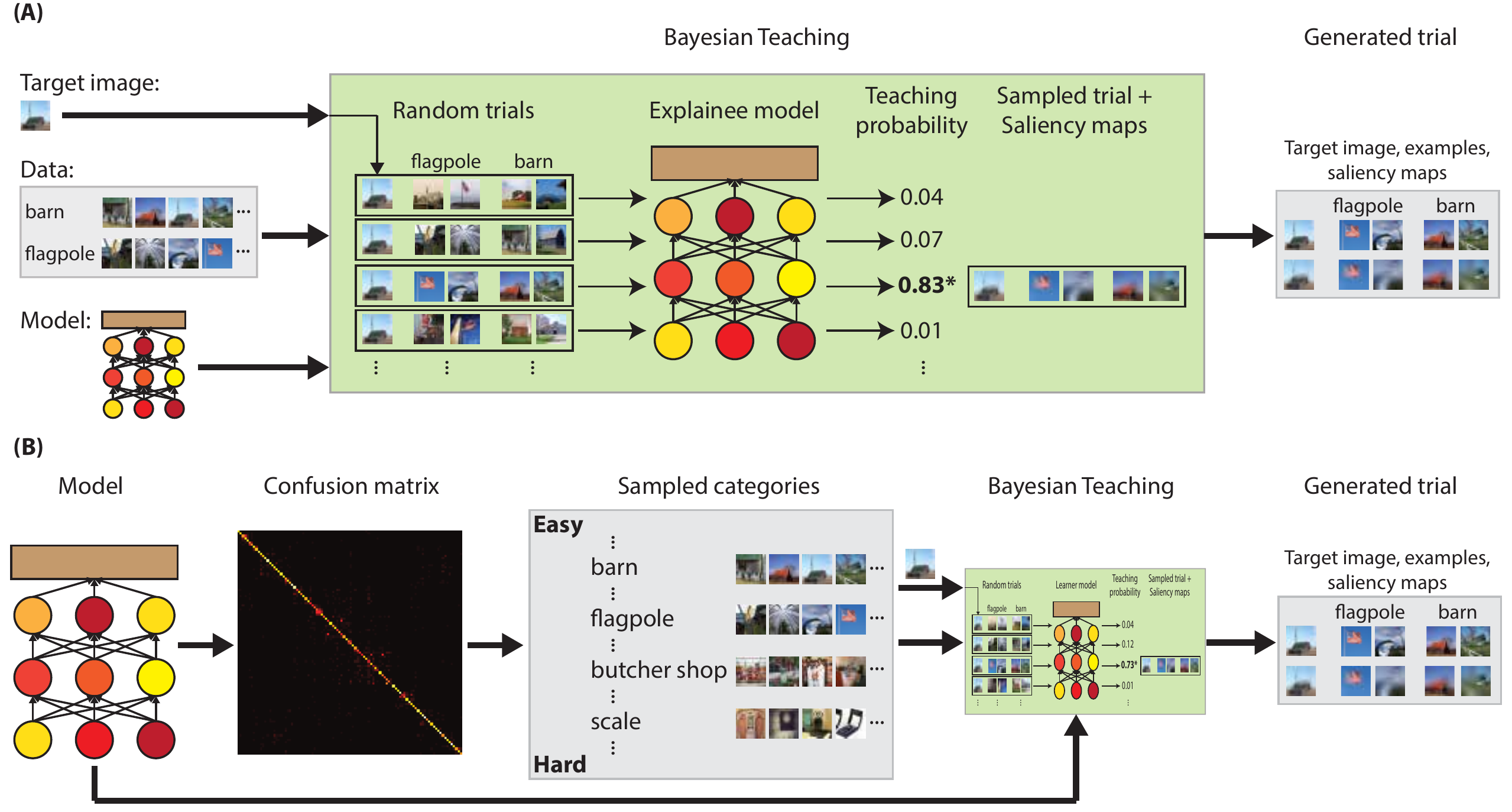}
  \caption{Flowchart of trial generation. \textbf{(A)}. Selection of examples and saliency maps with Bayesian Teaching. The inputs to Bayesian Teaching are: the model to be explained, data sets from two categories, and a target image that belongs to one of the two categories. The green box depicts the inner working of Bayesian Teaching. Random image pairs are selected from each of the input categories. Along with the target image, two sets of image pairs, one set from each category, are selected at random to form a trial. The \sy{explainee model}, which is set to have the same architecture as the input model, takes in a large number of random trials to produce the \sy{simulated explainee fidelity} (unnormalized teaching probabilities according to Equation~\ref{eq:learner_model}). Here, a trial with high \sy{fidelity} (probability) is selected, exemplifying the trial generation process in the [\textsc{helpful}] condition. Saliency maps are generated for the target image and the four selected examples using Equation~\ref{eq:expected_map_MC}. The final output is a set of ten images: a target image, two examples selected from each of the two input categories, and the saliency maps of the above five images. \textbf{(B)}. Trial generation steps peripheral to Bayesian Teaching. Our model to be explained is a ResNet-50 trained on \textit{ImageNet 1K}. A confusion matrix on the 1000 ImageNet categories was computed using the model. Using the confusion matrix, we sampled 25 categories where the model has high accuracy (the ``Easy" categories), 25 categories where the model has low accuracy (the ``Hard" categories), and the categories that are most confusable with the above 50 categories. To generate a trial, we select at random two categories from the 100 candidates mentioned above as well as a target image that belongs to one of the two selected categories. The model, the target image, and the data associated with the two categories are fed into Bayesian Teaching to produce a trial. See Methods for the full details. \sy{This figure was created using Adobe Illustrator CS6 (v.~16.0.0) \cite{adobeillustrator}}.} 
  \label{fig:flowchart}
\end{figure}

From this subset where both the image and the top model prediction belonged to our 100 categories, we randomly sampled 50 images where the model prediction matched the ground truth labels and 100 images for which the model predictions did not match the ground-truth labels. For the 50 trials with correctly classified target images, the two classification options participants could choose from were the correct model-predicted category and one of the two most confusable categories (out of our 100 selected categories). Which one of the two most confusable categories was presented were selected randomly for each trial. For the 100 incorrectly classified trials the two classification options were simply the ground-truth category and the incorrect model prediction. This procedure resulted in a total of 83 unique categories used in the experiment (Supplementary Table T1). This number is smaller than 100 because not all confusable categories are unique and not all categories were kept during the random sampling. Figure~\ref{fig:flowchart} depicts the trial generating process. The pairs of categories used in the experiments are listed in Supplementary Table T2.

\subsection*{Experimental design}
At the beginning of the experiment, participants were told that a robot has been trained to classify images but sometimes makes mistakes. They were asked to help by guessing how the robot will classify images. On each trial, a target image was displayed along with information about two categories, and the participants were asked to perform the 2AFC task by choosing which of the two categories they think the robot would classify the target image as.

The experimental conditions determined what information was presented during each trial and varied three dimensions: labels, examples, and saliency maps. Figure~\ref{fig:exp} shows a trial in the experimental condition with all the elements---labels, examples, and saliency maps---and describes how the conditions impact what elements are presented. More precisely, the conditions are characterized by five binary features: informative or generic labels, with or without examples, helpful or random examples (if present), with or without saliency maps, and blur or jet saliency maps (if present). The structured column and row labels of Table~\ref{tab:participants} show the naming conventions for the different conditions in terms of these features. Below, we provide more details on the conditions.

\textbf{Specific or generic labels:} Conditions with informative or generic labels are referred to as [\textsc{specific labels}] and [\textsc{generic labels}], respectively. In the [\textsc{specific labels}] conditions, the English labels of the two categories (e.g., ``Flagpole" and ``Barn" in Figure~\ref{fig:exp}) are given. In the [\textsc{generic labels}] conditions, the two categories are named ``Category A" and ``Category B."

\textbf{With or without examples:} Conditions with and without examples are referred to as [\textsc{examples}] and [\textsc{no examples}], respectively. In the [\textsc{examples}] conditions, two examples are sampled from each of the two categories to represent the category. Thus, five images---one target image and four example images---are on display in each trial in these conditions. In the [\textsc{no examples}] conditions, only the target image is shown.

\textbf{Helpful or random examples:} Conditions with the helpful examples and random examples are referred to as [\textsc{helpful}] and [\textsc{random}], respectively. The selection of the examples are based on the \textit{\sy{simulated explainee fidelity}}, which is the numerator of the Bayesian Teaching probability, $f_L\sy{(\cdot)}$. The \sy{simulated explainee fidelity} characterizes the probability that the four examples will lead a "\sy{explainee model}" to classify the target image as the ResNet-50-PLDA model would. The Bayesian Teaching probability and its numerator $f_L\sy{(\cdot)}$ are rigorously defined in Equations~\ref{eq:BT} and \ref{eq:learner_model}, respectively, in the subsection below called ``Selection of examples with Bayesian Teaching." In the [\textsc{helpful}] conditions, the four examples are chosen such that $f_L\sy{(\cdot)}>0.8$. In the [\textsc{random}] conditions, the four examples are chosen such that the $f_L\sy{(\cdot)}$ values across the 150 trials are uniformly distributed over the five bins that evenly partition the [0,1] interval.

\textbf{With or without saliency maps:} Conditions with and without saliency maps are referred to as [\textsc{map}] and [\textsc{no map}], respectively. A saliency map is an image mask that shows the contribution of each pixel to the model's classification decision. Details on the generation of the saliency maps are provided in the subsection below called ``Selection of saliency maps with Bayesian Teaching." In the [\textsc{map}] conditions, a saliency map is shown for every image displayed. In the [\textsc{no map}] conditions, no saliency map is shown.

\textbf{Blur or jet saliency maps:} Conditions with the blur saliency maps and jet saliency maps are referred to as [\textsc{blur}] and [\textsc{jet}], respectively. The two types of map differ only in the rendering of the mask but not in the generation of the mask. The jet saliency map renders the importance of each pixel by colors following the jet color map. In order of decreasing importance, the jet color map goes from red to green to blue. The jet color map, overlaid on an image with some level of transparency, is one of the most commonly used renderings of saliency maps. Two disadvantages of jet saliency maps are that the colors of the map can interfere with the colors of the image and that the unimportant regions remain visible to the user and can attract involuntary visual attention. For these reasons, we created the [\textsc{blur}] conditions in which the saliency maps are rendered by blurring the image. Furthermore, blurring is a more naturalistic visual effect than any color map masking because our visual system constantly experiences a large difference in visual acuity between our fovea and peripheral vision. The implementation details of both renderings are provided in the subsection below on saliency map selection.

\textbf{Naming convention:} As shown in Table~\ref{tab:participants}, not all combinations of the five binary features are allowed. Conditions with generic labels and no examples are not tested because that would make the 2AFC task a game of pure guessing. Furthermore, conditions without examples cannot be paired with helpful or random examples, and conditions without saliency maps cannot be paired with blur or jet maps. This leaves a total of 15 experimental conditions.

The naming convention for the conditions is based on filter queries using the database structure presented in Table~\ref{tab:participants}. To give a few examples: [\textsc{helpful}] refers to the aggregate of the six conditions in columns 2 and 4;  [\textsc{map}] refers to the aggregate of the 10 conditions in rows 2 and 3; [\textsc{helpful}] \& [\textsc{blur}] refers to the aggregate of the two conditions in row 2 column 2 and row 2 column 4; and [\textsc{helpful}] \& [\textsc{blur}] \& [\textsc{specific labels}] refers to the one condition in row 2 column 2.

\subsection*{Participants}
The study protocol was approved by Rutgers University IRB. All research was performed in accordance with the approved study protocol. An IRB-approved consent page was displayed before the experiment. Informed consent was obtained from all participants. The experiment began after the participants gave consent. 

656 participants (404 male, 249 female, 3 other) were recruited from Amazon Mechanical Turk and paid \$2.50 for completing the experiment, which took roughly 15 minutes to complete. The mean age of participants was 34.8 years (SD = 10.1), ranging from 18 to 72 years. The participants were randomly assigned to each condition, with the aim to obtain 36-40 participants per condition. 25 participants were excluded from analysis for completing the experiment too quickly (less than one second per trial), resulting in a final sample of 631 participants completing 150 trials each. The [\textsc{no examples}] conditions received twice the sample size of the other conditions, so that they would match the sample size of the [\textsc{examples}] conditions, which had two distinct versions ([\textsc{helpful}] and [\textsc{random}]). Table~\ref{tab:participants} shows the number participants in each of the 15 conditions. 

The number of participants in every condition is shown in Table~\ref{tab:participants}. All participants in the [\textsc{helpful}] conditions experienced the same set of 150 trials, i.e., the same 150 combinations of target image, category pairs, and example images, but in randomized order. All participants in the [\textsc{random}] conditions experienced another set of 150 trials, also in randomized order. All the category pairs used are listed in Supplementary Table T2. Participants in the [\textsc{no examples}] condition experienced one of these two sets of trials, selected at random. Note that because there are no examples but only English labels in the [\textsc{no examples}] conditions, the two sets of trials are functionally equivalent.

\subsection*{Selection of examples with Bayesian Teaching}
The goal of Bayesian Teaching is to select small subsets of the training data such that the inference made by a \sy{explainee} model using this small subset will be similar to the inference made by a target model using the entire training data. For this study, the target model is the ResNet-50-PLDA model trained on the 100 selected categories as described earlier. The inference task is to classify the target image among the 100 categories. The inference task of the \sy{explainee} model is the 2AFC image classification task presented in each trial. For the \sy{explainee} model, we search for an ideal-observer model   \cite{geisler2003ideal,geisler2011contributions} that would capture the participant's inference in the 2AFC task. A good candidate is the ResNet-50-PLDA because it is trained on human-labeled data and achieves high accuracy on predicting humans’ labelling behavior. This means that the target model and \sy{explainee} model share the same parameters (the ResNet-50 weights and PLDA parameters mentioned after Equation~\ref{eq:plda_predictive}), and the use of Bayesian Teaching is focused on explaining the image classification inference based on roughly 100K training examples, i.e., all the training data in the 100 selected categories, with only four training examples, i.e., those selected to be displayed on each trial of the experiments in the [\textsc{examples}] conditions.

We introduce some notation to define the Bayesian Teaching probability formally. The two categories that define the 2AFC task in each trial consist of the predicted category of the ResNet-50-PLDA model and an alternative category, which we denote by $\targetlabel$ and $\contrastivelabel$, respectively. The two examples sampled from the model-predicted category are denoted by $\ts^{\targetlabel}$, and the two sampled from the alternative category are denoted by $\ts^{\contrastivelabel}$. Let the \sy{explainee} model be denoted by $f_L$ and the target image be denoted by $\targetdatum$. The Bayesian Teaching probability, $P_T$, is defined as the probability that the selected examples, $\ts^{\targetlabel}$ and $\ts^{\contrastivelabel}$, will lead the \sy{explainee} model to classify the target image as the target model would. Mathematically, this probability can be expressed using Bayes' rule as: 
\begin{linenomath}
\begin{align}
P_T(\ts^{\targetlabel}, \ts^{\contrastivelabel}
  \mid \targetlabel, \targetdatum)
&=\frac{
    f_L(\targetlabel \mid \ts^{\targetlabel},
        \ts^{\contrastivelabel}, \targetdatum)
    }
    {
  \sum_{(\ts^{\targetlabel}, \ts^{\contrastivelabel})'
      \in \Omega}
    f_L (\targetlabel \mid 
         (\ts^{\targetlabel}, \ts^{\contrastivelabel})',
         \targetdatum)
  }.
\label{eq:BT}
\end{align}
\end{linenomath}
The sum in the denominator is over all possible candidate sets of the four examples. The set of all candidate sets is denoted by $\Omega$. Equation~\ref{eq:BT} assumes a uniform prior over $\Omega$ so that the prior terms in the numerator and denominator cancel out. Technically, $\Omega$ is the Cartesian product of all possible pairings of images in the category $\targetlabel$ with all possible pairings of images in the category $\contrastivelabel$, which is on the order of $10^{11}$ for the dataset in use. Our goal here is to select $\ts^{\targetlabel}$ and $\ts^{\contrastivelabel}$ such that $f_L\sy{(\cdot)}$, the numerator of Equation~\ref{eq:BT}, would provide good coverage of the full range of [0,1]. This would ensure the existence of valid examples for both the [\textsc{random}] and [\textsc{helpful}] conditions. We found that the full range can usually be covered by forming a Cartesian product of 1000 random pairings from each category ($10^6$ combinations). In general, given a target value of $f_L\sy{(\cdot)}$, one could use genetic algorithm \cite{back1996evolutionary} or other types of discrete optimization method to select the examples. To sample in proportion to $P_T$, one could use Markov Chain Monte Carlo and variational inference techniques \cite{haario2006dram,maclaurin2015firefly,chen2018learning}. These optimization and advanced inference methods would also be more efficient in the case that more than a few examples for each category is desired.

Using Bayes' rule again, we express the \sy{explainee} model's inference of the target image's label given the target image and examples, the numerator in Equation~\ref{eq:BT}, as
\begin{linenomath}
\begin{align}
f_L(\targetlabel
    \mid \ts^{\targetlabel},
    \ts^{\contrastivelabel},
    \targetdatum)
&=\frac{f(\targetdatum \mid \ts^{\targetlabel})
  }{f(\targetdatum \mid \ts^{\targetlabel})
    + f(\targetdatum \mid \ts^{\contrastivelabel})
  },
\label{eq:learner_model}
\end{align}
\end{linenomath}
where $f(\targetdatum \mid \ts^{k})$ is the probability that the target image, $\targetdatum$, belongs to the category from which the two example images, $\ts^{k}$, are sampled. Under the PLDA model, one can write this probability in closed form as a normal distribution \cite{vong2018}: 
\begin{linenomath}
\begin{align}
f(\targetdatum \mid \ts^k)
&=\mathcal{N}\left(u^* \,\middle\vert\, \frac{\Psi}{2\Psi + \text{I}} (u_1^k + u_2^k),\, \frac{\Psi}{2\Psi + \text{I}} + \text{I}\right).
\label{eq:plda_predictive}
\end{align}
\end{linenomath}
Here, $u$ is an image transformed in two steps. First, the image is passed through ResNet-50 and transformed into a feature vector; then, this feature vector undergoes an affine transformation with shift vector $\textbf{m}$ and rotation and scaling matrix $A$ to become $u$. Thus, in Equation~\ref{eq:plda_predictive}, $u^*$ is a transformed target image, and $(u_1^k, u_2^k)$ are a pair of transformed examples sampled from category $k$. The quantities $\textbf{m}$ and $A$ in the second transformation and the $\Psi$ in Equation~\ref{eq:plda_predictive} are parameters of the PLDA model obtained by training on the images in the 100 selected categories. The precise definitions of these parameters and the training procedure are presented in Figure 2 in Ioffe's PLDA paper \cite{ioffe2006probabilistic}. 

To summarize this subsection, Equation~\ref{eq:BT} defines the Bayesian Teaching probability, and Equation~\ref{eq:learner_model} defines its numerator (simulated \sy{explainee} \sy{fidelity}), $f_L\sy{(\cdot)}$, used to select examples in the [\textsc{examples}] conditions. A high $f_L\sy{(\cdot)}$ means that the selected examples will lead the model of the explainee to classify the target image as the category predicted by the model to be explained with high probability. Conversely, a low $f_L\sy{(\cdot)}$ means that the selected examples will lead the \sy{explainee} model to classify the target image as the other category in the 2AFC with high probability. Note that $f_L\sy{(\cdot)}$ is trial specific, as this probability is a function of the target image, $d^*$, the model predicted label of the target image, $y^*$, and the four examples, $(\ts^{y^*}, \ts^{y})$, which precisely define a trial.

\subsection*{Selection of saliency maps with Bayesian Teaching}
A saliency map is an image mask that shows how important each pixel of the image is to the model's inference. In the [\textsc{map}] conditions, we generate a saliency map for every image displayed. To generate a saliency map, one needs to specify a model, an inference task, and a definition of importance. We used ResNet-50 as the model and the classification of an image into the 1000 categories in \textit{ImageNet 1K} as the inference task. Using the Bayesian Teaching framework, we define importance to be the probability that a mask, $m$, will lead the model to predict the image, $\datum$, to be in category, $y$, when the mask is applied to the image. This is expressed by Bayes' rule as
\begin{linenomath}
\begin{align}
  Q_T(m \mid y, \datum) = \frac{g_L(y \mid \datum, m) p(m)}
  {\int_{\Omega_M} g_L(y \mid \datum, m) p(m)}.
  \label{eq:BT_map}
\end{align}
\end{linenomath}
Here, $g_L(y \mid \datum, m)$ is probability that the ResNet-50 model will predict the $\datum$ masked by $m$ to be $y$; $p(m)$ is the prior probability of $m$; and $\Omega_M = [0, 1]^{W \times H}$ is the space of all possible masks on an image with $W\times H$ pixels. The prior probability distribution $p(m)$ on $m$ is a sigmoid-function squashed Gaussian process prior.

Instead of sampling the saliency maps directly from Equation~\ref{eq:BT_map}, we find the expected saliency map for each image by Monte Carlo integration:
\begin{linenomath}
\begin{align}
\text{E}[M \mid y, \datum]
&=\int_{\Omega_M} m\ Q_T(m \mid y, \datum) \nonumber\\
&\approx \frac{\sum_{i=1}^N m_i\ g_L(y \mid \datum, m_i)}
    {\sum_{i=1}^N g_L(y \mid \datum, m_i)},
    \label{eq:expected_map_MC}
\end{align}
\end{linenomath}
where $m_i$ are samples from the prior distribution $p(m)$, and $N=1000$ is the number of Monte Carlo samples used. To see why an expected map is desirable, imagine the following case. Suppose that an image contains 7 goldfish and its category is ``goldfish." In this case, a mask that reveals any one of the goldfish will have a high $Q_T$ value. However, it is more desirable that the mask would reveal all the goldfish in the image. The expectation provides this by averaging the masks appropriately weighted by their $Q_T$ values. 

Now, we describe the step-by-step procedures for generating the saliency map for an image, $d$. First, $d$ is resized to be 224-by-224 pixels, which is the size displayed in the experiments (Figure~\ref{fig:exp}). A set of 1000 2D functions are sampled from a 2D Gaussian process (GP) with an overall variance of $100$, a constant mean of $-100$, and a radial-basis-function kernel with length scale 22.4 pixels in both dimensions. The sampled functions are evaluated on a 224-by-224 grid, and the function values are mostly in the range of $[-500,300]$. A sigmoid function, $1 / (1 + \exp(-x))$, is applied to the sampled functions to transform each of the function values, $x$, to be within the range $[0,1]$. This results in 1000 masks. The mean of the GP controls how many effective zeros there are in the mask, and the variance of the GP determines how fast neighboring pixel values in the mask change from zero to one. The 1000 masks are the $m_i$'s in Equation~\ref{eq:expected_map_MC}. We produce 1000 masked images by element-wise multiplying the image $d$ with each of the masks. The term $g_L(y \mid \datum, m_i)$ is the ResNet-50's predictive probability that the $i^\textrm{th}$ masked image is in category $y$. Having obtained these predictive probabilities from ResNet-50, we average the 1000 masks according to Equation~\ref{eq:expected_map_MC} to produce the saliency map of image $d$. If $d$ is a target image, the $y$ used to generate the saliency map is the ResNet-50-PLDA model's prediction. If $d$ is an example, the $y$ is the category from which the example is sampled. 

In the [\textsc{jet}] conditions, the saliency maps are rendered in the Matplotlib package with the ``jet" colormap and an alpha value of 0.4 and overlaid on the images (see Figure~\ref{fig:exp}; images at the bottom row). In the [\textsc{blur}] conditions, a saliency map is rendered by blurring the image for which it is generated (Figure~\ref{fig:exp}; images in the middle row). To generate the blur, each pixel value of a saliency map, $z$, is assigned a blurring window width, $w(z) = \textrm{ceil}(30/(1 + \exp(20z-10)))$. The $j^\textrm{th}$ pixel value of the rendered saliency map is the average pixel value of a patch of the original image, where the patch is $w$-by-$w$ in size and centered on the $j^\textrm{th}$ pixel of the original image. If the $j^\textrm{th}$ pixel is close to an edge of the image, the patch becomes rectangular, and the average is over whichever pixel values that are inside the $w$-by-$w$ sized window.

To conclude this subsection, we make a few final remarks. First, a PLDA layer is unnecessary in the generation of saliency maps because the ResNet-50 model is capable of generating the probabilities $g_L(y \mid \datum, m)$ in Equation~\ref{eq:BT_map}. In contrast, the ResNet-50 model cannot be used directly to generate the probabilities $f_L(\targetlabel \mid \ts^{\targetlabel}, \ts^{\contrastivelabel}, \targetdatum)$ in Equation~\ref{eq:BT}. Second, while the 2AFC task may be suitable for generating a saliency map for the target image, it cannot be used to generate saliency maps for the examples. This is the main reason that here we used the inference task of the image classification task that the ResNet-50 model is trained on. Lastly, Equation~\ref{eq:expected_map_MC} is the same as Equation 5 in the RISE approach introduced by Petsiuk, Das, and Saenko \cite{petsiuk2018rise}, which presents a state-of-the art method for generating saliency maps. Our implementation and their implementation differ only in the way the individual masks are sampled. In our implementation, we sampled functions from a GP prior and turned them into masks by applying a sigmoid function. In \cite{petsiuk2018rise}, random binary matrices are first sampled and subsequently up-sampled to the desired mask size through bilinear interpolation. The expectation is done in the same way.

\subsection*{Familiarity coding}
In addition to the splits by conditions presented in Table~\ref{tab:participants}, analysis also rely on scores of human familiarity with the image categories. The familiarity of each of the [\textsc{helpful}] and [\textsc{random}] trials was manually coded by 7 raters. Each rater was asked to code the trial as ``familiar" if they thought they could correctly match the category labels to the images presented in that trial, and ``unfamiliar" otherwise. A familiarity score for each pairing of categories was then constructed by assigning each raters judgements as 1 for familiar and 0 for unfamiliar, and computing the mean across raters. The 300 trials across the [\textsc{helpful}] and [\textsc{random}] conditions resulted in 167 unique category pairings (counting the ordering of target versus other category), and their familiarity scores are presented in Supplementary Table T2.

\subsection*{Statistical analysis}
Whenever we report testing how well participants predict the model classifications (\sy{fidelity}), or how often their judgements correspond to the image ground truth (accuracy) we used hierarchical logistic regressions with random intercepts per participants and fixed effects for the remaining terms. For analyses of sensitivity and specificity analyses, we still used a logistic regression framework but only included trials corresponding to true positives and false negatives, or true negatives and false positives, respectively. Sensitivity captures how well participants predict trials when the AI is correct, and specificity capture how well participants predict trials when the model is wrong.

To illustrate, \textit{Bayesian Teaching improves \sy{fidelity}} used the following equations on the full set of trials, and on a subset of the trials to capture sensitivity and specificity respectively:
\begin{linenomath}
\begin{align*}
\mathrm{Pr(\sy{ParticipantChoice}_{i}=\sy{AIChoice}_{i})} &=\mathrm{logit}^{-1} (\alpha_{j[i]} + \beta_{1}  \mathrm{ExplanationCondition}_{i} + \epsilon_{i}), \;\mathrm{for}\; i = 1, \dots, I.    \\
\mathrm{logit}^{-1} (x) &= \frac{\exp(x) }{1 + \exp(x)} \\
\alpha_{j} &\sim N(U_{j}, \sigma^{2}_{\alpha}), \;\mathrm{for}\; j = 1, \dots, J.
\end{align*}
\end{linenomath}
where \sy{the agreement between a participant's choice and the AI classifier's choice} is a binary variable coded as 1 when participant correctly predict the AI classification and 0 otherwise, i is the observation index, j is the participant index. ExplanationCondition is a binary dummy variable coded as 1 if participants experienced heatmaps and helpful examples and 0 if they did not experience any explanations.

For the \textit{Participants prefer helpful examples} section we compared three hierarchical logistic models: (A) An intercept only model that treated intercepts as nested within participants (B) an intercept only model that treated intercepts as nested within participants and conditions, (C) Model two, with an added fixed effect for the familiarity score. We then compared the negative log-likelihoods of these models to determine which best accounted for the observed data.

We evaluated whether \textit{Bayesian Teaching can lead participants to both correct and incorrect inference} by predicting \sy{fidelity} in the conditions containing examples by fitting three nested models:
\begin{linenomath}
\begin{align*}
\mathrm{Pr(\sy{ParticipantChoice}_{i}=\sy{AIChoice}_{i})} &=\mathrm{logit}^{-1} (\alpha_{j[i]} + \beta_{1}  \mathrm{\sy{CategoryAccuracy}}_{i} + \epsilon_{i}), \;\mathrm{for}\; i = 1, \dots, I. \\
\mathrm{Pr(\sy{participant\;choice}_{i}=\sy{AI\;choice}_{i}} &=\mathrm{logit}^{-1} (\dots + \beta_{2}  \mathrm{\sy{SimExplaineeFidelity}}_{i} + \epsilon_{i}), \;\mathrm{for}\; i = 1, \dots, I. \\
\mathrm{Pr(\sy{participant\;choice}_{i}=\sy{AI\;choice}_{i})} &=\mathrm{logit}^{-1} (\dots + \beta_{3}  \mathrm{ModelCorrectness}_{i}+ \\\nonumber
&\beta_{4}  \mathrm{ModelCorrectness}_{i} \mathrm{\sy{CategoryAccuracy}}_{i}  + \\\nonumber &\beta_{5}  \mathrm{ModelCorrectness}_{i}\mathrm{\sy{SimExplaineeFidelity}}_{i} + \epsilon_{i}), \;\mathrm{for}\; i = 1, \dots, I.
\end{align*}
\end{linenomath}
where \sy{SimExplaineeFidelity} is the expected probability that the participant pick the same response as the target model, conditional on seeing the examples, \sy{CategoryAccuracy, is the average classification accuracy of the target ResNet-50 model for the target category,} and ModelCorrectness is a dummy variable coding if ResNet made a correct classification on this particular trial. We then compared the negative log likelihoods of these three models, and reported the coefficients of the best-fitting model (the interaction model).

In the \textit{Bayesian Teaching improves \sy{fidelity} through belief-mitigation} section we fitted four logistic hierarchical regression models to the full data. These models shared the following form:
\begin{linenomath}
\begin{align}
\mathrm{Pr(Y_{i}=1)} &=\mathrm{logit}^{-1} (\alpha_{j[i]} + \beta_{1}  \mathrm{\sy{FamiliarityScore}}_{i} + \beta_{2} \mathrm{\sy{CategoryAccuracy}}_{i} + \\\nonumber &\beta_{3} \mathrm{Examples}_{i} + \beta_{4} \mathrm{MAP}_{i} + \beta_{5} \mathrm{Labels}_{i} + \epsilon_{i}), \;\mathrm{for}\; i = 1, \dots, I. \label{eq:bt_main}
\end{align}
\end{linenomath}
where \sy{FamiliarityScore} is a proportion of raters who rated the trial categories as familiar, Examples, MAP and Labels where dummy variables that captured whether examples were shown, whether heatmaps were shown and whether category labels were informative or not, respectively.

These four models were distinguished based on whether the AI was correct or not and whether Y corresponded to whether the participant judgement matched the ground truth or matched the AI's judgement. We fitted similar models to the [\textsc{examples}] trials only, with the only difference being that the Examples term, that previously had captured whether examples were present was replaced with a dummy variable that captured whether the examples presented were helpful or not. Finally, we fitted two more models predicting \sy{fidelity} from the full data. These are similar to Equation~6, but added two additional interaction terms:
\begin{linenomath}
\begin{align*}
\mathrm{Pr(Y_{i}=1)} &=\mathrm{logit}^{-1} (\dots + \beta_{6}  \mathrm{MAP}_{i}\mathrm{\sy{FamiliarityScore}}_{i} + \beta_{7}\mathrm{Examples}_{i} \mathrm{sy{FamiliarityScore}}_{i} + \epsilon_{i}), \\&\mathrm{for}\; i = 1,\; \dots, I.
\end{align*}
\end{linenomath}
Coefficient tables for these models can be found in Supplementary Tables T3. All hierarchical logistic regression models were fitted using the lme4 package (1.1-23) \cite{bates2007lme4} in R version 4.0.3 \cite{r_core}, Figures were created in ggplot 2 version 3.3.2 \cite{ggplot}.

\section*{Data availability}
Raw experimental data and analysis code will be deposited at \url{https://github.com/CoDaS-Lab/XAI-BT-SR} upon publication.

\section*{Acknowledgments}
This material is based on research sponsored by the Air Force Research Laboratory and DARPA under agreement number FA8750-17-2-0146 to P.S. and S.Y. The U.S. Government is authorized to reproduce and distribute reprints for Governmental purposes notwithstanding any copyright notation thereon.

This work was also supported by DoD grant 72531RTREP, NSF SMA-1640816, NSF MRI 1828528 to P.S. The methods described herein are covered under Provisional Application No. 62/774,976.

\section*{Author contributions}
S.C.-H.Y., W.K.V., R.B.S, and P.S. conceived and designed the experiments. W.K.V. and S.C.-H.Y. conducted the experiments. R.B.S. and S.C.-H.Y. prepared materials for the experiments. T.F., W.K.V., and S.C.-H.Y. analyzed the data. S.C.-H.Y., T.F., W.K.V., R.B.S, and P.S. wrote the paper.

\section*{Competing interests}
The authors declare no competing interests.

\bibliography{layout4arxiv}
\bibliographystyle{unsrt}

\newpage
\section*{\Large \centering Supplementary information: Mitigating belief projection in explainable artificial intelligence via Bayesian Teaching} 

\section*{Supplementary Table T1}
ImageNet categories used in the experiment. The 83 categories and their corresponding \sy{category accuracy} are given in this table. The accuracy scores are computed on the test set of \textit{ImageNet 1K} over the 100 selected categories described in Methods. See separate csv file. 

\section*{Supplementary Table T2}
The table lists all 167 unique pairs of categories used in the experiment along with each pair's familiarity score. See separate csv file.

\section*{Supplementary Tables T3}
Coefficient tables for the 15 regression models reported in the main text. See separate excel file.

\newpage
\section*{Supplementary Discussion D1: Participants prefer helpful examples}
\subsection*{Methods}
To test the subjective preference for helpful versus unhelpful or random examples, we use a different task. In a trial of this task, we presented a target image, its category, and two sets of example pairs for that category. The participants were asked to select which pair they think influenced the AI's classification more. Note that this experiment is different from the 2AFC experiment described previously in that there is only one category and the decision is between teaching sets.

Helpful examples are chosen to be the teaching examples for the target category in trials where $f_L\sy{(\cdot)}>0.8$. Likewise, unhelpful examples are chosen to be the examples for the target category in trials where $f_L\sy{(\cdot)}<0.2$. On average, these examples are expected to be helpful or detrimental regardless of what the other category is; thus, they can be approximated as examples that aim to maximize or minimize the marginal teaching probability. We extracted 67 target images that have both helpful examples and unhelpful examples. Given a target image's category, random examples are simply random samples from the training images in \textit{ImageNet 1K} that are not the target image or the helpful examples.


80 participants (25 male, 54 female, 1 other) were recruited from Amazon Mechanical Turk and paid \$1.00 for completing the experiment, which took roughly 5 minutes to complete. The participants were randomly assigned to one of the two conditions (helpful-vs-unhelpful and helpful-vs-random) with 40 in each condition. The mean age of participants was 36.7 years (SD = 10.5), ranging from 16 to 68 years. 6 participants were excluded from analysis for completing the experiment too quickly (less than one second per trial), resulting in a final sample of 74 participants.

The study protocol was approved by Rutgers University IRB. All research was performed in accordance with the approved study protocol. An IRB-approved consent page was displayed before the experiment. Informed consent was obtained from all participants. The experiment began after the participants gave consent.

\subsection*{Results}
We wanted to evaluate whether participants preferred informative to uninformative and misleading examples. To test this, we ran a second study where participants chose between helpful examples versus random examples (n=37) or versus unheplful examples (n=37). The helpfulness of the examples is determined by Bayesian Teaching. The helpful examples are chosen to best represent the target category by maximizing the marginal teaching probability; random examples are uniformly randomly sampled from the target category; and unhelpful examples are chosen to mislead the learner to infer any other category by minimizing the marginal teaching probability. The marginal teaching probability is the probability that a set of examples will lead the explainee model to infer the target category compared to any other category in a 2AFC task (see Methods for more details).

Participants showed a small but reliable preference for helpful relative to random examples (53.05\% [95\% CI = 51.08\% - 55.01\%], z=3.03, p = .002) and a substantial preference for helpful versus to unhelpful examples (64.14\% [95\% CI = 61.68\% - 66.59\%], z=10.95, p $<$ .0001). These two conditions were reliably different (\textchi2 = 36.94, p $<$ .0001), implying that the Bayesian Teacher is not only capable of selecting helpful examples, but can also select examples that are actively confusing (see Figure ~\ref{fig:preference}). As this pattern of preferences matches our predictions as stated in the introduction, a natural next steps is to evaluate whether these preferences are particularly pronounced for unfamiliar examples, as hypothesised. We found that participants were more likely to prefer helpful examples when the choice categories were unfamiliar to them (\textbeta = -0.57(0.08), z = -7.02, p $<$ .0001), irrespective of whether helpful examples were contrasted with random or unhelpful examples.

\renewcommand{\thefigure}{D1-1}
\begin{figure}[h]
  \centering
  \includegraphics[width=\columnwidth]{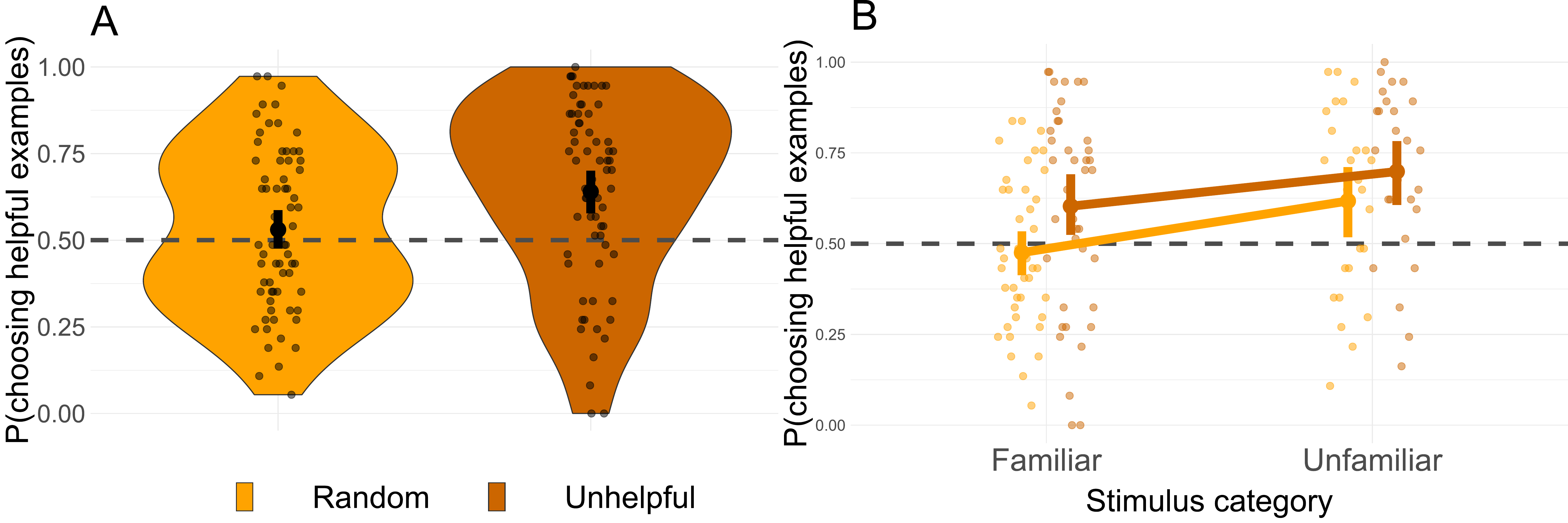}
  \caption{\sy{Helpful examples are preferred to unhelpful and random examples, especially for unfamiliar categories}. \textbf{(A)}. The probability that a participant chose helpful examples over random (37 participants; 2479 observations), or unhelpful examples (37 participants; 2479 observations) respectively. \textbf{(B)}. The less familiar participants are with the stimulus categories, the more they prefer helpful examples. Familiarity ratings were continuous in the analyses reported in the main text, but are dichotomized here for visual clarity. Each transparent point represents the average probability across participants for one specific stimulus pair. Solid points represent the mean across all stimulus pairs. Error bars signify 95\% bootstrapped confidence intervals. \sy{This figure was created using the ggplot2 package (v. 3.3.2) \cite{ggplot} in R (v. 4.0.3) \cite{r_core}.}} 
  \label{fig:preference}
\end{figure}

\newpage
\section*{Supplementary Discussion D2: Analysing MAP conditions separately}
In the main text we combined the two \textsc{map}] conditions in our analyses. To show that this decision did not meaningfully impact our conclusions we repeat the same analyses with [\textsc{jet}] and [\textsc{blur}] as separate predictors here. We ran hierarchical logistic regressions on the complete dataset predicting \sy{the fidelity between the participant predictions of the classifications of the AI model and its actual classifications based the explanatory interventions} ([\textsc{specific-labels}] vs [\textsc{generic-labels}], [\textsc{blur}] vs [\textsc{jet}]  vs [\textsc{no map}], and [\textsc{examples}] vs [\textsc{no examples}], while controlling for \sy{category accuracy} and familiarity ratings.

[\textsc{blur}] improves \sy{fidelity} when the \sy{AI classifier} is wrong (\textbeta = 0.43(0.03), z = 12.27, p $<$ .0001), as do [\textsc{jet}] (\textbeta = 0.43(0.03), z = 12.34, p $<$ .0001). However, the \sy{saliency maps} reduce \sy{fidelity} (to a lesser extent) when the \sy{AI classifier} is correct, both for [\textsc{blur}]  (\textbeta = -0.49(0.08), z = -6.06, p $<$ .0001) and for [\textsc{jet}]  (\textbeta = -0.62(0.08), z = -7.76, p $<$ .0001), see Figure~\ref{fig:supplement}. In both cases, the saliency maps reduced the first order-accuracy of the participants. [\textsc{blur}] AI correct: \textbeta = -0.49(0.08), z = -6.05, p $<$ .0001, AI wrong: \textbeta = -0.43(0.03), z = -12.27, p $<$ .0001; [\textsc{jet}] AI correct: \textbeta = -0.62(0.08), z = -7.76, p $<$ .0001, AI wrong: \textbeta = -0.43(0.03), z = -12.34, p $<$ .0001. This reduction in first-order accuracy means that participants were less likely to believe that the AI judgements matched the ground truth of the image. This in turn implies that the saliency maps encourage participants to consider that the AI might be mistaken. This interpretation assumes that participants know the ground truth for most of the trials, which seems plausible \sy{given typical human classification accuracy on the ImageNet dataset \cite{ILSVRC15}}. 

The familiarity ratings capture the ease of the discrimination task in that they are higher for trials involving categories that humans are familiar with. We can use these ratings to further explore whether participants project their own beliefs onto the AI. Specifically, if humans use their first-order classifications to model the AI, familiarity should positively correlate with \sy{fidelity} when the AI is correct, but negatively correlate with \sy{fidelity} when the AI is wrong. This is indeed what we find: participants are more likely to accurately predict AI classifications when they are familiar with the item categories and the AI is correct (\textbeta = 1.10(0.04), z = 29.28, p $<$ .0001), but they are less likely to correctly predict AI errors (\textbeta = -0.92(0.02), z = -42.82, p $<$ .0001). In other words, participants are more likely to assume that the AI gets it right for trials that they themselves find easy.

Previously we showed that saliency maps improved prediction accuracy on trials when the AI was wrong. We suggested that this might be explained by saliency maps helping participants distinguish between their first-order judgements of the ground truth and their \sy{fidelity when} predicting the model classification. This can be evaluated directly by testing whether the impact of the familiarity ratings on classification accuracy are attenuated by the saliency maps (see Figure~\ref{fig:supplement}). In other words, if participants are more likely to predict that the AI is correct on trials that they themselves find easy, and the saliency maps work by helping people realise that the AI use different decision-processes, the saliency maps should make participants more willing to consider that the AI is wrong for trials they themselves find easy. This is what we find, see Figure~\ref{fig:supplement}). Specifically, the presence of [\textsc{blur}] maps reduces the positive impact of familiarity on \sy{fidelity} when the AI is correct (\textbeta = -0.61(0.09), z = -6.67, p $<$ .0001) and the same is true for [\textsc{jet}] maps (\textbeta = -0.44(0.09), z = -4.87, p $<$ .0001). Conversely, saliency maps reduce the negative impact of familiarity on \sy{fidelity} when the AI is wrong, for both [\textsc{blur}] (\textbeta = 0.74(0.05), z = 13.95, p $<$ .0001) and [\textsc{jet}] (\textbeta = 0.67(0.05), z = 12.73, p $<$ .0001). Collectively these results suggest that the presence of saliency maps help participants model the AI as an agent with distinct beliefs that may conflict with their own.

\renewcommand{\thefigure}{D2-1}
\begin{figure}[h!]
  \centering
  \includegraphics[width=\columnwidth]{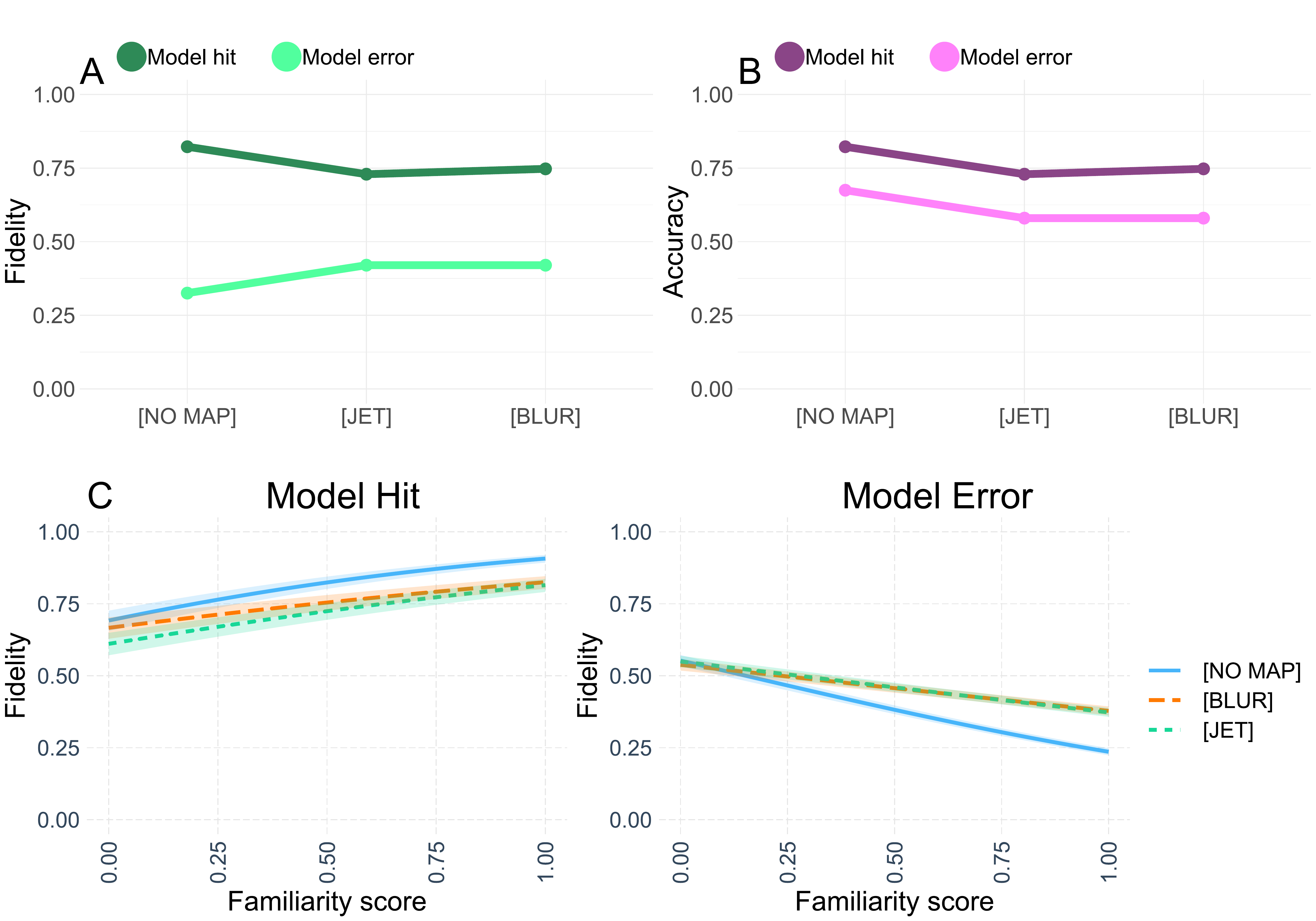}
  \caption{\sy{Saliency maps} improve human \sy{fidelity when} identifying model errors, and reduce \sy{fidelity when} identifying model hits, irrespective of \sy{the presentation format of the \sy{saliency maps}}. All subplots are based on the entire data set, comparing [\textsc{jet}] and [\textsc{blur}] and [\textsc{no map}] conditions (631 participants; 94,582 observations). \textbf{(A)}. The saliency maps improve \sy{fidelity} for trials when the AI is wrong but reduce \sy{fidelity} when the AI is correct. \textbf{(B)}. The saliency maps make people less likely to classify the target image to align with the ground truth, independent of AI accuracy. Together, A \& B imply that the saliency maps help people to consider that the AI might make mistakes. \textbf{(C)} Saliency maps decrease the impact of familiarity on participant judgements. For model hits this leads to decreased \sy{fidelity}, whereas for model errors it leads to improved \sy{fidelity}.  This pattern provides further evidence that the saliency maps work by shifting participants away from using their first-order judgments to model the AI classifications. Collectively these figures suggest that [\textsc{jet}] and [\textsc{blur}] have very similar impacts on participant judgements. Errorbars represent 95\% bootstrapped confidence intervals. All point estimates have confidence intervals, though some are too narrow to see clearly. Shaded areas represent analytic 95\% confidence intervals. \sy{This figure was created using the ggplot2 package (v. 3.3.2) \cite{ggplot} in R (v. 4.0.3) \cite{r_core}.}} 
  \label{fig:supplement}
\end{figure}

\newpage
\section*{Supplementary Figure F1: The relationship between \sy{category accuracy} and participant \sy{fidelity} in the control condition}
Focusing exclusively on the control trials, we see that \sy{category accuracy} is positively associated with human \sy{fidelity} when the AI is wrong (\textbeta = 0.81(0.11), z = 6.95, p $<$ .0001), but even more so when the AI is correct (\textbeta = 0.92(0.23), z = 3.96, p $<$ .0001).  
\renewcommand{\thefigure}{F1}
\begin{figure}[ht!]
  \centering
  \includegraphics[width=\columnwidth]{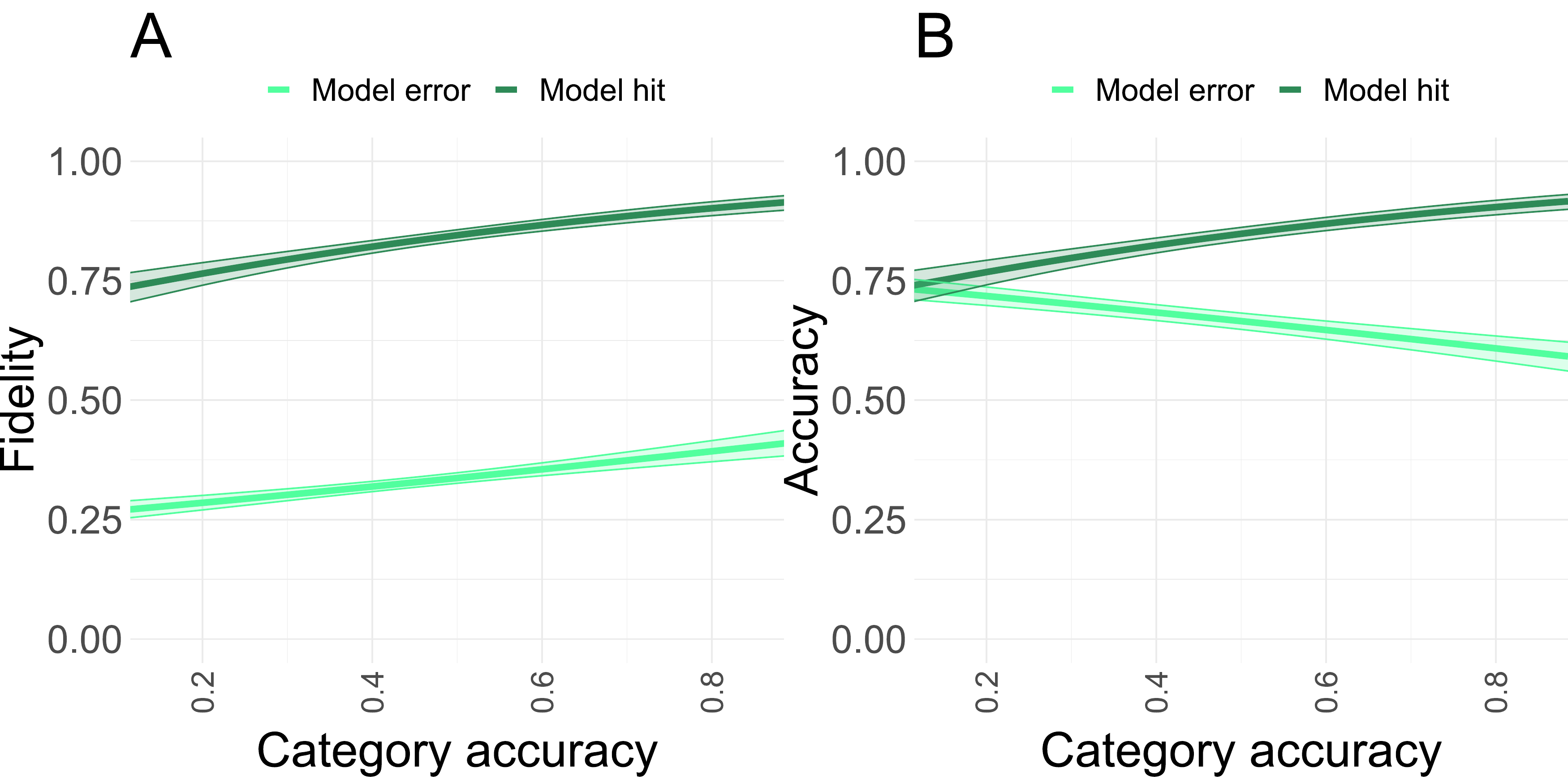}
  \caption{\sy{category accuracy versus participant \sy{fidelity} in the control condition.} \textbf{(A)} \sy{category accuracy} is positively associated with human \sy{fidelity} during the control condition both for trials when the AI is correct and when the AI is wrong. However, the base rate human \sy{fidelity} is much higher when the AI is correct. \textbf{(B)} The probability that the participant judgement corresponds to the ground truth is positively associated with \sy{category accuracy} when the model is correct, but negatively associated with \sy{category accuracy} when the model is wrong. Both subplots are based on the control trials only (76 participants; 11,394 observations).  Shaded areas represent analytic 95\% confidence intervals. \sy{This figure was created using the ggplot2 package (v. 3.3.2) \cite{ggplot} in R (v. 4.0.3) \cite{r_core}.}} 
  \label{fig:supplement_teacher}
\end{figure}

\end{document}